%% file: main_camera_ready.tex
\documentclass{article}

\usepackage[final]{corl_2024} 
\usepackage[export]{adjustbox}
\usepackage{amsfonts}
\usepackage{amsmath}
\usepackage{amssymb}
\usepackage{authblk}
\usepackage{booktabs}
\usepackage{caption}
\usepackage{colortbl}
\usepackage{enumitem}
\usepackage{float}
\usepackage{graphicx}
\usepackage{lipsum}
\usepackage{wrapfig}
\usepackage{multirow}
\usepackage{multicol}
\usepackage{nicefrac}
\usepackage{pifont}
\usepackage{subcaption}
\usepackage{url}
\usepackage{wrapfig}
\usepackage{bbding}
\usepackage{scalerel}
\usepackage{tikz}   
\usepackage{xcolor}

\newcommand{\flame}{\scalerel*{\includegraphics{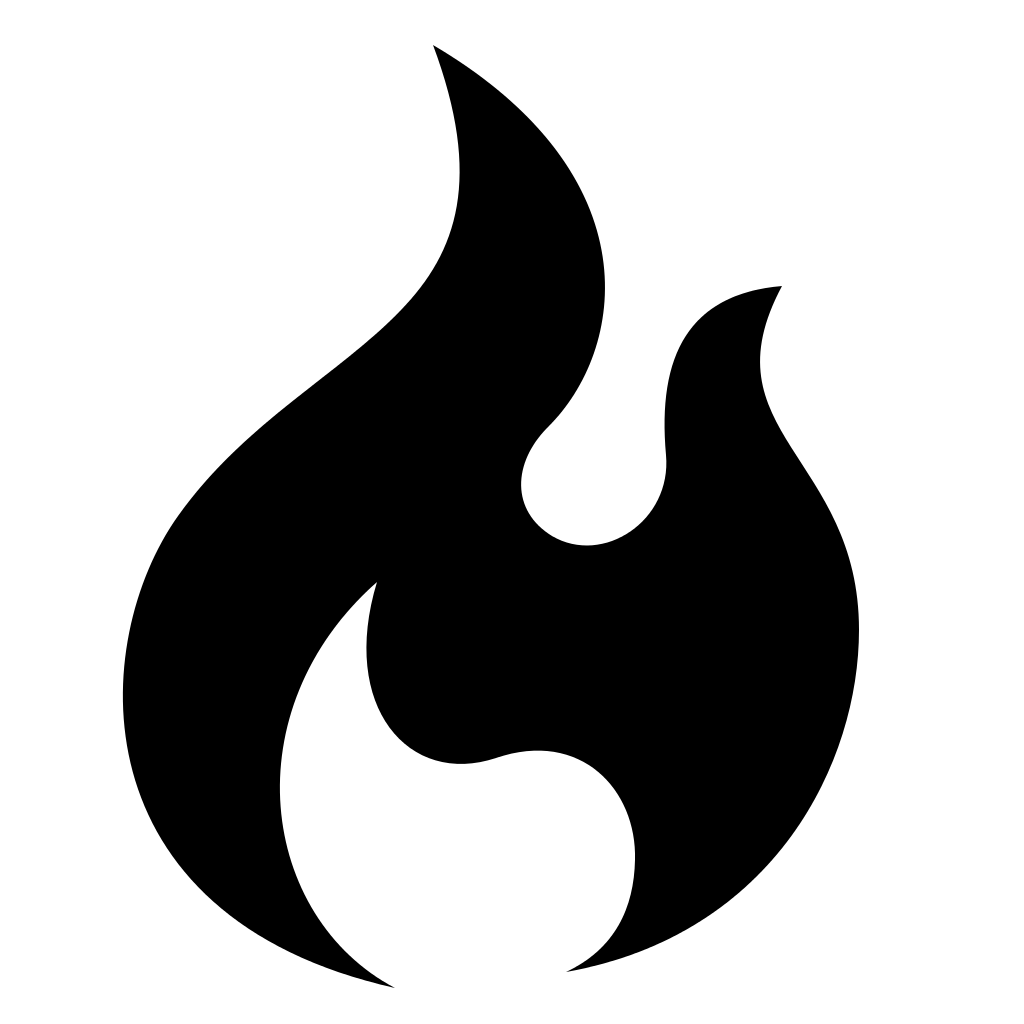}}{X}}
\let\oldsnowflake\Snowflake
\renewcommand{\Snowflake}{\raisebox{-2pt}{\oldsnowflake}}
\makeatletter
\renewcommand\AB@affilsepx{\quad \protect\Affilfont}
\makeatother
\newcommand*\numcircledmod[1]{\raisebox{.5pt}{\textcircled{\raisebox{-.9pt} {#1}}}}

\title{Theia: Distilling Diverse Vision Foundation Models for Robot Learning\vspace{-15pt}}

\usepackage{authblk}

\renewcommand{\Affilfont}{\normalfont\small\centering}

\author[1,2]{Jinghuan Shang}
\author[1]{Karl Schmeckpeper}
\author[1]{Brandon B. May}
\author[1]{Maria Vittoria Minniti}
\author[1]{\newline Tarik Kelestemur}
\author[1]{David Watkins}
\author[1]{Laura Herlant}
\vspace{-8pt}
\affil[ ]{\hfill $^\text{1}$The AI Institute}
\affil[2]{Stony Brook University\hfill \break}

\affil[ ]{\texttt{\{jshang, kschmeckpeper, bmay, mminniti, tkelestemur, dwatkins, lherlant\}@theaiinstitute.com}\vspace{-30pt}}

\begin{document}
\maketitle
\setlist[itemize]{leftmargin=15pt}

\begin{abstract}
Vision-based robot policy learning, which maps visual inputs to actions, necessitates a holistic understanding of diverse visual tasks beyond single-task needs like classification or segmentation. Inspired by this, we introduce Theia, a vision foundation model for robot learning that distills multiple off-the-shelf vision foundation models trained on varied vision tasks. Theia's rich visual representations encode diverse visual knowledge, enhancing downstream robot learning. Extensive experiments demonstrate that Theia outperforms its teacher models and prior robot learning models using less training data and smaller model sizes. Additionally, we quantify the quality of pre-trained visual representations and hypothesize that higher entropy in feature norm distributions leads to improved robot learning performance. Code, models, and demo are available \href{https://theia.theaiinstitute.com}{here}.
\end{abstract}


\input{figures/performance_over_model_size}
\vspace{-10pt}
\section{Introduction}
\vspace{-2pt}
Visual understanding, i.e., the process of abstracting high-dimensional visual signals like images and videos, includes many different sub-problems, from depth prediction~\cite{yang2024depth} and vision-language correspondence~\cite{radford2021learning,liu2024visual}, to tasks ranging from coarse to fine granularity such as classification~\cite{he2016deep,dosovitskiy2021an} and object grounding~\cite{caron2021emerging,oquab2023dinov2,locatello2020object}, as well as tasks defined along spatial and temporal axes like segmentation~\cite{long2015fully,kirillov2023segment} and tracking~\cite{yang2023track}. 
Given this diversity, a long-standing effort in the vision community has been to develop models tailored to one or a few specific types of visual understanding tasks. 
In recent years, several models~\cite{he2016deep,dosovitskiy2021an,kirillov2023segment,radford2021learning,liu2024visual,kingma2013auto,caron2020unsupervised,caron2021emerging,he2022masked,zhou2022ibot,kirillov2023segment,oquab2023dinov2,schuhmann2022laion} have achieved remarkable generalizability to unseen domains and new tasks, and are commonly referred to as vision foundation models (VFMs).

Vision-based robot policy learning, which learns action policies from visual inputs, requires strong and diverse visual comprehension. These policies involve many implicit vision tasks such as object recognition and semantic grounding, where off-the-shelf VFMs corresponding to some well-defined tasks can be easily found, but there is no single model for all of vision tasks. 
Studies have shown that off-the-shelf VFMs such as CLIP~\cite{radford2021learning} usually under-perform relative to visual representation models tailored for specific tasks in robot learning~\cite{nair2022r3m,ma2022vip,xiao2022masked,majumdar2024we,dasari2023unbiased}. 
This fact reveals a gap between the needs of robot learning and the limited visual understanding capabilities of any individual VFM. 
Our motivation is different than all prior works on learning visual representation models for robotics, where those works focused primarily on improving training data ~\cite{nair2022r3m, dasari2023unbiased, majumdar2024we}, designing objective functions~\cite{nair2022r3m, ma2022vip}, and directly taking advances of vision architectures~\cite{xiao2022masked}. We uniquely focus on improving the visual representation from the angle of solving multiple implicit visual understanding tasks, which will benefit the downstream robot learning.

In this work, we propose combining multiple large VFMs into a single, smaller model for robot learning that leverages diverse visual understanding abilities from VFMs. We achieve this via knowledge distillation~\cite{hinton2015distilling}. 
Unlike conventional distillation from a larger model to a smaller model on the same task, we distill VFMs tailored for varied vision tasks to improve visual representation for robot learning, which is an \textit{unseen} task for VFMs.

We introduce Theia, a robot vision foundation model that simultaneously distills off-the-shelf VFMs such as CLIP~\cite{radford2021learning}, DINOv2~\cite{oquab2023dinov2}, and ViT~\cite{dosovitskiy2021an}. 
Compared to off-the-shelf VFMs~\cite{radford2021learning,dosovitskiy2021an,oquab2023dinov2} and prior works~\cite{xiao2022masked,majumdar2024we}, Theia offers both better pre-trained visual representations for higher downstream robot learning performance and reduced computational costs. Furthermore, training Theia only requires ImageNet~\cite{jia2009imagenet} and about 150 GPU hours, in contrast to prior works which necessitate substantially more compute~\cite{xiao2022masked,majumdar2024we,nair2022r3m, ma2022vip}. To understand what makes a good visual representation for robot learning, we observe multiple factors that relate to the performance of downstream robot learning tasks. We hypothesize that higher entropy in representation norms~\cite{darcet2024visionregister} correlates with improved robot learning performance.

In summary, our contributions are:
\vspace{-7pt}
\begin{itemize}
\setlength\itemsep{-2pt}
    \item We introduce Theia, a model that combines knowledge from multiple VFMs into a single, smaller model using knowledge distillation with low training cost. 
    \item Through extensive simulated and real-world experiments, we confirm that Theia's visual representations lead to better downstream robot learning with improved computational efficiency.
    \item We identify key factors relevant to robot learning performance, such as model size, the use of spatial tokens, and the entropy of representation norms, offering valuable insights for guiding future research on optimizing visual representations for robot learning.
\end{itemize}

\vspace{-15pt}
\section{Related Work}
\vspace{-5pt}
\subsection{Visual Representations for Robot Learning}
\vspace{-5pt}
Visual representations are important for vision-based robot policies to parse high-dimensional visual signals. Visual representation learning can happen at different stages, including pre-training~\cite{nair2022r3m, ma2022vip, majumdar2024we, dasari2023unbiased}, joint-learning with robot tasks~\cite{laskin2020reinforcement,yarats2020image,laskin2020curl,li2022does}, or a combination of both using either trainable or frozen visual representations~\cite{shah21rrl,wang2022vrl}. Off-the-shelf visual encoders~\cite{he2016deep,dosovitskiy2021an,radford2021learning,li2024llara,kim24openvla} can also provide visual representations for robot learning. Additionally, an important factor in training visual representations is the choice of data. ImageNet~\cite{jia2009imagenet}, as suggested by~\citet{dasari2023unbiased},  is a particularly effective pre-training dataset, while video datasets~\cite{grauman2022ego4d, damen2018scaling} are also widely used. Training objectives and auxiliary tasks for visual representation learning vary, including data augmentation~\cite{laskin2020reinforcement,yarats2020image}, prediction tasks~\cite{li2022does}, contrastive learning~\cite{laskin2020curl,sermanet2018time,shang2021self}, and self-supervised learning~\cite{xiao2022masked,majumdar2024we}. Specifically, to handle invariance and equivariance in visual observations, inductive biases~\cite{shang2022learning,shi2024composing} and constraints~\cite{wang22equi,chen2023equidiff} can be introduced into neural networks to improve visual representation quality. Unlike prior works, we build a robot vision foundation model from the perspective of merging the visual understanding abilities of VFMs via distillation. Concurrent work OpenVLA~\cite{kim24openvla} uses pre-trained SigLIP~\cite{zhai2023sigmoid} and DINOv2~\cite{oquab2023dinov2} encoders, which suggests the benefit of diverse visual understanding in robot tasks.

\vspace{-7pt}
\subsection{Vision Foundation Models}
\vspace{-5pt}
Vision foundation models trained on large-scale data exhibit strong task-specific performance and transferability to unseen domains and new tasks. 
VFMs can focus on single tasks, multiple tasks, or remain task-agnostic at the pre-training stage. 
For example, ViT~\cite{dosovitskiy2021an}, DeiT~\cite{touvron2021training}, and ConvNeXt~\cite{liu2022convnet} are designed for image classification, SAM~\cite{kirillov2023segment} for semantic segmentation, and Depth-Anything~\cite{yang2024depth} for depth prediction. When combined with Large Language Models (LLMs)~\cite{alayrac2022flamingo,liu2024visual,chen2023minigpt,xu2023u}, these models can solve multiple visual tasks including referring segmentation, visual question answering (VQA), and image editing. Task-agnostic models like the DINO series~\cite{caron2021emerging,oquab2023dinov2} are trained through self-distillation, while CLIP~\cite{radford2021learning} is trained by aligning image-text pairs. Given the strong generalization capability of VFMs, robot learning should also benefit from leveraging the latent representations of pre-trained VFMs. which is the primary motivation for our research.

\vspace{-7pt}
\subsection{Knowledge Distillation in Vision Models}
\vspace{-5pt}
Knowledge distillation~\cite{hinton2015distilling} compresses the knowledge from one or more larger models into a single smaller model. To leverage VFMs, which are usually computationally intensive, several studies have explored distilling them into more compact models. For example, some works distill a single VFM like SAM~\cite{kirillov2023segment} into smaller variants~\cite{chen2023slimsam, xiong2023efficientsam, zhang2024efficientvit}. There are also relevant works on combining two models: SAM-CLIP~\cite{wang2023samclip} merges SAM~\cite{kirillov2023segment} and CLIP~\cite{radford2021learning} into one model, while ViLD~\cite{gu2021open} and GroundingDINO~\cite{liu2023grounding} combine vision and language models to perform object detection. RADIO~\cite{ranzinger2023amradio} is an agglomerative model that simultaneously distills CLIP, SAM, and DINOv2. These studies show that combined models can improve performance on downstream tasks or enable new applications. Similarly, in this work, we investigate whether combining VFMs will benefit robot learning. Notably, RADIO~\cite{ranzinger2023amradio} is the closest approach to ours. The key differences between Theia and RADIO~\cite{ranzinger2023amradio} are that (1) We aim to use our representation for robot learning tasks that none of the VFMs have covered in their pretraining tasks, (2) we distill only spatial tokens rather than both spatial and \texttt{[CLS]} tokens, (3) we choose a different set of teacher models, and (4) we show how each teacher model contributes to robot learning performance.

\vspace{-7pt}
\section{Method}
\vspace{-7pt}
\paragraph{Overview.}
\input{figures/theia_main_figure}
We introduce Theia, a framework that distills the knowledge of multiple VFMs into a smaller model, producing rich spatial representations for downstream vision-based robot learning.
Figure~\ref{fig:theia_main_figure} shows the overall design of Theia.
Our model comprises a visual encoder (backbone) and a set of feature translators for distillation.
Note that only the visual encoder is used to produce latent representations for downstream robot learning tasks. 

\vspace{-5pt}
\paragraph{Architecture.}
\vspace{-3pt}
Given an input image $\mathbf{x}$, the visual encoder $f(\cdot)$ produces a rich representation $\mathbf{z}=f(\mathbf{x})$ (called the Theia-representation), which is utilized for downstream robot learning tasks. 
We focus on backbone models that are smaller than typical VFMs, specifically using ViT-Tiny, Small, and Base~\cite{dosovitskiy2021an,touvron2021training}\footnote{For simplicity, we denote DeiT-Tiny and DeiT-Small~\cite{touvron2021training} by ViT-Tiny and ViT-Small. DeiT was introduced by~\citet{touvron2021training} and the original ViT~\cite{dosovitskiy2021an} does not have Tiny and Small models.}
motivated by the limited computing resources on robotic systems. We use Theia-Tiny (Theia-T), Theia-Small (Theia-S) and Theia-Base (Theia-B) to refer to Theia models using a vision transformer backbone with the corresponding size. To train the Theia-representation, we perform feature distillation with the help of feature translators, which will be described below. 

\vspace{-5pt}
\subsection{Rich Spatial Representation}
\vspace{-3pt}
The Theia-representation is a set of encoded tokens corresponding to input image patches. We choose spatial tokens because spatially-dense representations are the foundation for diverse visual understanding, as evidenced by the powerful per-patch features in DINOv2~\cite{oquab2023dinov2}. Therefore, we aim to distill all spatial tokens and leave the \texttt{[CLS]} token untouched. 

\vspace{-5pt}
\paragraph{Feature Translators.} 
\vspace{-3pt}
Our goal is to supervise Theia-representations with teacher representations from various VFMs.
We extract teacher representations $h_i(\mathbf{x})$ of VFMs at the last layer for CLIP~\cite{radford2021learning}, ViT~\cite{dosovitskiy2021an} and DINOv2~\cite{oquab2023dinov2}, or before the decoders for SAM~\cite{kirillov2023segment} and Depth-Anything~\cite{yang2024depth}.
Since a single representation cannot be learned to match all the teacher representations directly, feature translators $g_i(\cdot)$ are used to map the Theia-representation, $\mathbf{z}$, to each teacher representation. 
Feature translators are shallow CNNs to ensure that knowledge is distilled primarily into Theia's visual encoder. Details are available in the Appendix. 

\vspace{-5pt}
\subsection{Training}
\vspace{-3pt}
\paragraph{Distillation Objective.} Our training objective is matching the outputs of the feature translators with their corresponding teacher VFM representations. To achieve this, we use a combination of cosine and smooth-L1 losses~\cite{ranzinger2023amradio} to match each pair of predicted and ground truth representations for the same image, taking their weighted average. Formally, our loss is
\begin{equation}
    \mathcal{L}(\mathbf{x};\theta) = \sum_i^M\alpha_i\bigl(\beta\mathcal{L}_{cos}(g_i(f(\mathbf{x})), h_i(\mathbf{x})) + (1-\beta)\mathcal{L}_{smooth-L1}(g_i(f(\mathbf{x})), h_i(\mathbf{x}))\bigr),
\end{equation}
where $\mathbf{x}$ is the input image, $M$ is the number of teacher VFMs, $\alpha_i$ is the loss weight for each teacher, and $\beta$ is the weight for balancing cosine loss and smooth-L1 loss respectively. In general, we set $\alpha_i=1/M$ such that the loss weights each teacher equally.
We empirically set $\beta=0.9$~\cite{ranzinger2023amradio}. 

\vspace{-5pt}
\paragraph{Feature Normalization.} To properly accommodate the different scales of teacher representations, we first perform a normalization step. This helps us scale the loss of different teacher features evenly and avoid biasing (collapsing) to a teacher model with extremely larger norms. 
We perform the normalization on the teacher representations over each latent dimension, where mean and variance are calculated from all ImageNet training samples. Details are in Appendix~\ref{sec:appendix_theia_model_architecture}.

\vspace{-5pt}
\paragraph{Dataset.} We train our model on the ImageNet~\cite{jia2009imagenet} training set for 50 epochs. We opt to use ImageNet because of its greater diversity compared to human videos~\cite{grauman2022ego4d, goyal2017ssv2,shan2020100doh,damen2018epickitchens} and robot datasets~\cite{kalashnikov2018scalable,dasari2019robonet,ebert2021bridge,lynch2023interactive,padalkar2023oxe} within the same number of images. This diversity has been experimentally shown to improve visual representation learning~\cite{dasari2023unbiased}.

\vspace{-7pt}
\section{Experiments}
\input{figures/robots_combined}
\vspace{-5pt}
\subsection{Benchmark and Settings}
\vspace{-3pt}
To evaluate pre-trained visual representations, we use simulation tasks in CortexBench~\cite{majumdar2024we}, which combines MuJoCo tasks (Adroit~\cite{Kumar2016adroit,rajeswaran2017learningadroit}, DeepMind Control Suite (DMC)~\cite{tassa2018dmc}, and MetaWorld~\cite{yu2020metaworld}), Habitat~\cite{savva2019habitat1,szot2021habitat2} tasks (ImageNav~\cite{mezghan2022memoryimagenav}, ObjectNav~\cite{yadav2023habitatobjnav}, and MobilePick~\cite{gu2022multirearrange}.), and Trifinger~\cite{majumdar2024we} tasks\footnote{Due to reproducibility issues, we are not able to evaluate Move Cube, ObjectNav, and MobilePick tasks\label{footnote:reproducibility_issues}}. ImageNav and MobilePick are reinforcement learning (RL) tasks and others are imitation learning (IL) tasks. We follow the experiment settings of~\citet{majumdar2024we} and report aggregated scores for rewards (DMC tasks) and success rates (all other tasks). For DMC tasks, raw rewards are divided by 10 to be in a scale consistent with the success rate. 
We also conduct real robot experiments, introduced in Section~\ref{sec:sec_real_world_robot_learning}. 
We use the same policy heads for the each type of representations (vector or spatial tokens). 
Full experimental details are available in the Appendix.

\vspace{-5pt}
\subsection{Simulation Results}\label{sec:simulation_results}
\vspace{-5pt}
We comprehensively evaluate Theia and baseline pre-trained models on the MuJoCo subset of CortexBench~\cite{majumdar2024we} to provide an overall assessment of pre-trained visual representations. We consider prior works on visual representations for robot learning, including R3M~\cite{nair2022r3m}, VIP~\cite{ma2022vip}, MVP~\cite{xiao2022masked}, and VC-1~\cite{majumdar2024we}, as well as agglomerative models for vision tasks RADIO and E-RADIO~\cite{ranzinger2023amradio}, and off-the-shelf vision foundation models ViT~\cite{dosovitskiy2021an}, DINOv2~\cite{oquab2023dinov2}, and CLIP~\cite{radford2021learning}. We also test a naive concatination of three VFMs (CLIP, ViT, and DINOv2) which are used to train Theia, referred as CDV. All pre-trained representations are frozen in this experiment.
Throughout this section, we answer the following questions:
\vspace{-5pt}
\begin{itemize}
\setlength\itemsep{-1pt}
    \item How does Theia perform compared to baselines?
    \item Which is more effective for visual representations: \texttt{[CLS]} or spatial tokens?
    \item How does robot learning performance scale with the size of the visual encoder?
\end{itemize}

\vspace{-10pt}
\paragraph{Theia Performance.}
\input{figures/mujoco_results}
As shown in Figure~\ref{fig:mujoco_results}, Theia outperforms all evaluated models, surpassing the performance of the best prior models, R3M and MVP, as well as agglomerative models for vision tasks RADIO and E-RADIO. We also tested a naive approach of using multiple VFMs (CLIP, DINOv2, and ViT) simultaneously by concatenating their spatial tokens channel-wise (CDV in Figure~\ref{fig:mujoco_results}), but this performed much worse than using just the individual VFMs. Theia models scale effectively from tiny to base sizes, with Theia-S and Theia-B being the only models to break scores of 80 on this subset of CortexBench, even though they use only a small fraction of the inference computation required by comparable models.
Theia's training is very efficient, using only the 1.2M images in ImageNet with a training time of about 150 GPU hours on NVIDIA H100 GPUs, compared to approximately 5M images used in prior works~\cite{nair2022r3m,ma2022vip,xiao2022masked,majumdar2024we} and 1B images used by RADIO~\cite{ranzinger2023amradio}. 

\vspace{-6pt}
\paragraph{Spatial Tokens vs. \texttt{[CLS]} Token.} We evaluate Transformer-based models (all models in Figure~\ref{fig:mujoco_results} except R3M and VIP) using either their \texttt{[CLS]} token or spatial tokens~\cite{dosovitskiy2021an} for downstream robot learning. To accommodate spatial tokens in the robot policy, we introduce extra shallow CNN layers at the input of the policy network, known as ``compression layer''~\cite{yadav2023ovrl,majumdar2024we}. Figure~\ref{fig:mujoco_results} shows the results of all models we evaluated, clearly showing that for Transformer-based models, providing spatial tokens is consistently better than using the \texttt{[CLS]} token for robot learning. This finding applies to both off-the-shelf VFMs and prior pre-trained representations, including MVP and VC-1. 

\vspace{-6pt}
\paragraph{Scaling with Model Size.} We observe that most models, including Theia, achieve better performance with larger sizes. The scaling effect is more obvious when using the \texttt{[CLS]} token, probably because the \texttt{[CLS]} token encodes less information, making the size of the feature vector more critical. Different models that use spatial tokens also scale at varying rates. VC-1 shows only minor improvements when scaling from base to large, while ViT shows much larger improvements when scaling from tiny to huge model.

\vspace{-6pt}
\paragraph{CortexBench Results.}
\input{tables/main_cortexbench_table}
We compare Theia-B against baselines over the CortexBench evaluation suite~\footref{footnote:reproducibility_issues}. The results in Table~\ref{tab:main_cortexbench_table} confirm that Theia-B outperforms all other models.

\vspace{-7pt}
\subsection{Real World Robot Learning}\label{sec:sec_real_world_robot_learning}
\vspace{-5pt}
\input{tables/real_robot_experiments}
Based on simulation performance, we test Theia-B and the best-performing baseline models: MVP-L~\cite{xiao2022masked}, R3M~\cite{nair2022r3m}, VC-1-L~\cite{majumdar2024we}, DINOv2-L~\cite{oquab2023dinov2}, ViT-H~\cite{dosovitskiy2021an}, and E-RADIO-L~\cite{ranzinger2023amradio} for evaluation on real-world tasks. We employ four tasks (Figure~\ref{fig:real_world_robot_learning_tasks}): \textit{Door Opening}, \textit{Pick-and-Place}, and \textit{Toy-Microwave Cooking} with a WidowX 250s arm, and \textit{Drawer Opening} with a Boston Dynamics Spot. We train behavioral cloning policies on top of visual representations using conventional policy networks composed of CNNs and MLPs in the WidowX setup and diffusion policy~\cite{chi2023diffusion} in the Spot setup. During testing, we vary the robot position for Door Opening and Drawer Opening, randomize the object position for Pick-and-Place, and randomize both the object positions and object types in Toy-Microwave Cooking. Full experimental settings, including details about the number of collected demonstrations, the interface used to collect them, and the policy architecture, are available in the Appendix (Sec.~\ref{sec:appendix_real_world_robot_learning}).

Table~\ref{tab:real_robot_experiments} shows the success rates on these real-world tasks.
Theia-B achieves the highest success rate across all tasks except Toy-Microwave Cooking. The results also highlight that the Theia-representation is useful for both conventional and diffusion-based policy heads, and for either  freezing or fine-tuning the visual representation. E-RADIO is the most competitive model in this setting amongst all models compared, likely due to its similar distillation of VFMs and much larger training dataset. VC-1 has high task variance, performing poorly on Door Opening but adequately on Pick-and-Place. ViT-H works much better when being fine-tuned but DINOv2 does not, which could be caused by some fundamental differences in VFMs.

\vspace{-7pt}
\subsection{Ablation Studies}
\vspace{-5pt}
\paragraph{Selection of Teacher Models.}\label{sec:experiments_combination}
\input{figures/combination_of_teacher_models}
Theia is motivated by distilling many vision foundation models, but not all the models can be effectively integrated into one model nor do they contribute equally to downstream robot learning. In these experiments, we empirically identify the most effective combination of teacher models among five candidate VFMs: CLIP (referred to as \textbf{C})~\cite{radford2021learning}, Depth-Anything (\textbf{De})~\cite{yang2024depth}, DINOv2 (\textbf{Di})~\cite{oquab2023dinov2}, SAM (\textbf{S})~\cite{kirillov2023segment}, and ViT (\textbf{V})~\cite{dosovitskiy2021an}. We select these candidates because they have been designed to perform well across various important vision tasks.

To select the best combination, we train Theia-T using different sets of teachers and evaluate the learned representation on the MuJoCo subset of tasks.
Figure~\ref{fig:combination_of_teacher_models} shows the results of different teacher combinations. We start with single VFM teachers and observe that \textbf{C} and \textbf{V} are the most beneficial. We then distill \textbf{All} teacher models altogether and discover that it performs better than four out of five single-teacher distillations, but performs worse than only distilling \textbf{V}, suggesting possible negative effects from some of the teacher models. 
We then remove each teacher model from \textbf{All} and distill the remaining four (\textbf{All}$-$\textbf{X}).
The results show that removing \textbf{V} and \textbf{C} from the teachers causes the most significant performance drops while taking out \textbf{S} and \textbf{De} leads to either similar or improved performance.
Given the (nearly) negative effects observed, we distill the rest of the three models (\textbf{CDiV}), which results in the best performance. We also try the  teacher combination that was performed in RADIO~\cite{ranzinger2023amradio} (\textbf{CDiS}), but it does not outperform \textbf{CDiV}.

\vspace{-3pt}
\paragraph{Model Design.}
\label{sec:experiments/ablations}
\input{tables/ablation_studies}
We conducted ablation studies on distillation loss choices and model architecture. We compared two kinds of distillation losses: MSE and Cos+L1 from~\citet{ranzinger2023amradio}. As shown in  Table~\ref{tab:distillation_losses}, we find that Cos+L1 performs better. For model architecture, we examined the effect of ``register tokens'' introduced by~\citet{darcet2024visionregister}. These extra tokens are introduced at the input without supervision and are discarded when using the representation. With the same amount of spatial tokens supervised by distillation, we varied the number of register tokens of between 0, 1, 4, and 8. Note the unsupervised \texttt{[CLS]} token is removed in zero register token case, and it counts for 1 register token in other cases. According to Table~\ref{tab:num_register_tokens}, we find that using 1 register token worked the best, while having no register tokens performed the worst. Compared to RADIO~\cite{ranzinger2023amradio}, which distills both \texttt{[CLS]} and spatial tokens, our use of the \texttt{[CLS]} token as a register token provides an advantage. We also evaluated distilling both \texttt{[CLS]} and spatial tokens in the Theia model, and the results shown in Table~\ref{tab:tokens_to_distill} confirm the advantage of exclusively distilling spatial tokens.

\vspace{-10pt}
\subsection{Qualitative Visualizations}\label{sec:qualitative_visualization}
\vspace{-3pt}
\input{figures/decoding_visualization}
We present qualitative visualizations to demonstrate how Theia-representations can be transformed into teacher representations through feature translators. Using Theia trained with all teachers (\textbf{CDeDiSV}, \textbf{All}), we applied PCA for visualizing predicted DINOv2~\cite{oquab2023dinov2} features, used the SAM~\cite{kirillov2023segment} decoder to produce segmentation results, and used the Depth-Anything~\cite{yang2024depth} head to produce estimated depth. Results are shown in Figure~\ref{fig:decoding_visualization} with more examples in the Appendix. The visualizations indicate that our predicted representations can be decoded by the original VFM and produce reasonable results. Encouragingly, we find that predicted depth maps from the Theia-predicted representation appear to be more accurate compared to the original Depth-Anything model, particularly in the stove-top area and at the microwave door. This shows the potential benefit of combining different visual understandings.

\vspace{-7pt}
\section{What Makes Visual Representations Good for Robot Learning?}
\vspace{-5pt}
\input{figures/feature_quality}
Traditionally, the quality of the pre-trained visual representations is evaluated through downstream robot learning like IL or RL. However, it is unclear why different visual representations lead to varying robot learning performance outcomes. In this section, we quantify the quality of visual representations and analyze how they correlate with downstream robot learning performance.  

\vspace{-5pt}
\paragraph{Feature Norm Distributions and Entropy.} \citet{darcet2024visionregister} analyzed the norm of spatial tokens in vision Transformers and found that high-norm outlier tokens are detrimental to vision task performance. Following this, we investigate whether a similar phenomenon arises in visual representations for robot learning. 
We inspect the feature norms of Theia with different teacher combinations and baseline models evaluated in Section~\ref{sec:simulation_results}, and their corresponding performance on the MuJoCo subset tasks. 
We sample 1\% of the MuJoCo task training set and calculate the L2-norm of each spatial token after encoding images. We measure the entropy of the feature norm distribution across all samples and patches per model and use it as a quantitative metric. To calculate the entropy of the distribution, we first discretize the distribution by a histogram. Then, we use the following formula to obtain the entropy $H = -\sum_i(p_i * \log(p_i))$, where $p_i$ is the probability of each discretized bin.

We confirm that similar outlier tokens also appear in VC-1 corresponding to the image patches that are not task-relevant, shown in the visualizations of feature norms on the right of Figure~\ref{fig:feature_quality}. In contrast, Theia has very few or no outlier tokens, and the tokens with higher norms are more task-relevant even though Theia-representations are not trained on these robot images. 
In our quantitative analysis (Figure~\ref{fig:feature_quality}, left), we divide the models into distilled and regular based on the observation that distilled models generally have higher entropy (fewer outliers, Figure 4(c) in~\citet{darcet2024visionregister}). We find that there is a strong correlation (R=0.943) between entropy and robot learning performance among regular models, and a high correlation (R=0.638) among distilled models. We hypothesize that spatial token representations with high entropy (better feature diversity) encode more information that aids policy learning, while less diverse representations (low entropy) may hinder it. 
In the Appendix~\ref{sec:analysis_visual_representations}, we discuss the results of other quantitative measurements, including feature similarity and PCA-explained variance ratios, where no strong correlations are found.

\vspace{-5pt}
\section{Conclusion}
\vspace{-5pt}
In this work, we introduced Theia, a novel robot vision foundation model specifically distilled from multiple VFMs to enhance robot learning tasks. Theia builds a rich visual representation from diverse VFM teachers, preserving spatial features to ensure detailed visual comprehension. Through extensive evaluations on CortexBench and in the real world, Theia consistently outperforms state-of-the-art models, including all prior models for robot learning, off-the-shelf VFMs, and similarly distilled models for vision tasks. Our results highlight the effectiveness of distilling multiple VFMs into a compact model for superior performance in a variety of robot learning scenarios. Furthermore, we answer a key question about what kinds of visual representations lead to better robot learning by finding a strong correlation between the entropy of feature norms and enhanced downstream performance, offering insights for future research on optimizing visual representations for robotics.



\acknowledgments{The authors thank Deva Ramanan for inspiring discussions on the analysis of trained models, Osman Dogan Yirmibesoglu for his help with the Spot grippers, and Joe St. Germain and Jien Cao for their assistance in setting up the WidowX robots. We also extend our gratitude to Erica Lin and Ahmet Gundogdu for their work on Spot’s policy learning infrastructure, as well as Nathan Williams and Brennan Vanderlaan for DevOps support.}


\bibliography{bibliography}  

\newpage
\appendix
\section{Theia Model Architecture}\label{sec:appendix_theia_model_architecture}
\paragraph{Backbone.} We use the DeiT-Tiny, DeiT-Small, and DeiT-Base models~\cite{touvron2021training} as our backbone architectures. We keep the \texttt{[CLS]} token in the model and in the forward pass, but there is no supervisory signal provided for it. As a result, the \texttt{[CLS]} token serves as a ``register token''~\cite{darcet2024visionregister}, which provides some benefits for learning high quality representations. We train Theia from scratch (no pre-trained DeiT~\cite{touvron2021training} weights are applied).
\paragraph{Feature Translators. } The feature translators are composed primarily of CNNs, with a linear layer appended at the end to match the teacher's representation dimension. Pure linear transforms might not be able to map Theia-representations to all three teacher representations well, resulting in a failure of learning (See Table~\ref{tab:ablation_feature_translator}). Thus, we use three CNN layers to account for the fact that each teacher model's representations are very different from one another. Details are listed in Table~\ref{tab:feature_translators}, where we show the architectural details of the translators used for our (student, teacher)-feature pairs.
\input{tables/feature_translators}

\paragraph{Feature Normalization.} Formally, the normalization is:
\begin{align}
    \Tilde{h(\mathbf{x}_j)}_c = \frac{h(\mathbf{x}_j)_c - \mu_c}{\sigma_c},~ \mu_c = \frac{1}{N}\sum_j{h(\mathbf{x}_j)_c},~ \sigma_c = \sqrt{\frac{\sum_j{(h(\mathbf{x}_j)_c-\mu)^2}}{N}},
\end{align}
where $c$ is the channel index for the feature, $j$ is the index of image sample, and $N$ is the number of samples in ImageNet.

\section{Training}
\input{tables/training_configuration}
We train Theia on 8 NVIDIA H100 GPUs. The main bottlenecks in training are the data transfer speed between devices and the GPU memory bandwidth to load large spatial feature tensors, for example, of size $1280 \mkern-5mu \times \mkern-5mu 16 \mkern-5mu \times \mkern-5mu 16$ for ViT-H and $256 \mkern-5mu \times \mkern-5mu 64 \mkern-5mu \times \mkern-5mu 64$ for SAM. We pre-compute the features from all teacher models instead of doing inference on the fly. This approach requires extra storage space to save all the features extracted from the VFMs, but significantly saves on training time and avoids loading models with high GPU memory usage during training, such as Depth-Anything or SAM (a batch size of 16 cannot fit into 80GB of GPU memory). All training configurations are listed in Table~\ref{tab:training_configuration}.

\paragraph{Teacher VFM Features.} We use the output representations at the last layer of ViT~\cite{dosovitskiy2021an}, CLIP~\cite{radford2021learning}, and DINOv2~\cite{oquab2023dinov2}. For SAM~\cite{kirillov2023segment}, we use its encoder output. For Depth-Anything~\cite{yang2024depth}, since it is initialized from DINOv2, we use the latent representation before the final convolution layer. When decoding SAM and Depth-Anything results from Theia-predicted representations, we send the predicted representations through the remaining layers of original models and obtain the output. 

\section{Additional Ablation Studies}
\input{tables/ablation_studies_appendix}

We conduct two additional ablation studies to verify design choices in the Theia model. The first is a comparison between the current CNN-based feature translator and a linear feature translator. In Table~\ref{tab:ablation_feature_translator}, we find that using a Linear feature translator leads to a significant performance drop. The second ablation studies whether Theia should be trained from scratch or be initialized using the pre-trained DeiT~\cite{touvron2021training} weights. In Table~\ref{tab:ablation_pretrained_backbone}, we find that using pre-trained weights improves the downstream performance. This could be interpreted as the positive effect of incorporating knowledge from an additional useful model into the distillation process. We would expect to see similar performance improvements as more informative models are included during training.

\section{Full Experimental Settings}
\input{tables/detail_baseline_data_objective_comparison}
\subsection{Baseline Models}
Theia and baseline models are trained on different sizes of datasets using different objectives. We organize these details in Table~\ref{tab:detail_baseline_data_objective_comparison} to provide a comprehensive comparison between them.

\subsection{CortexBench}
For all of our CortexBench experiments, we use the original setup~\cite{majumdar2024we}, except for a few modifications to produce more reliable results. The modifications include:
\begin{itemize}
    \item We increase the number of evaluation roll-outs from 10 (original) to 25 (ours) in DMC tasks. The mean scores reported are from a total of 75 runs (25 per seed x 3).
    \item We remove the noise added to the policy network output in the CortexBench code base. The noise causes minor performance degradation (about 1.0 on overall mean score for MuJoCo tasks) compared to the version without noise. 
    \item We modify the policy networks to take spatial feature inputs for MuJoCo and Trifinger tasks (details follow and are presented in Table~\ref{tab:policy_network}).
\end{itemize}
Note that prior models including R3M, VIP, MVP, and VC-1 are all re-run using the same settings in MuJoCo tasks for the purposes of making a fair comparison when evaluating against Theia.

\paragraph{Policy Networks.}
For MuJoCo and Trifinger tasks we utilize a three-layer MLP for vector-based representations, including ResNet models and Transformer models that use the \texttt{[CLS]} token. For models that generate spatial feature maps, such as Transformers using spatial tokens, we introduce a three-layer CNN before the MLP. For Habitat tasks, we exclusively benchmark models that produce spatial feature maps and adopt the same policy network as used by~\citet{majumdar2024we}. Details can be found in Table~\ref{tab:policy_network}.
\input{tables/policy_network}

For ImageNav task, we use the provided policy network from VC-1~\cite{majumdar2024we} without modification. The policy network is composed by a compression-layer (a CNN layer) to convert the spatial feature map into a vector representation, followed by a 2-layer LSTM. Details are available in Appendix A.2 in~\cite{majumdar2024we}.

\subsection{Real World Robot Learning}\label{sec:appendix_real_world_robot_learning}
\input{tables/experimental_details_table.tex}
\subsubsection{WidowX Arm Experiments}
\paragraph{WidowX Arm Setup.}
The robot used for these experiments is a 6-degree-of-freedom (DOF) WidowX 250s arm. The data collection and evaluation framework is based on~\cite{ebert2021bridge}. 

We train a behavior-cloning policy for each of the four evaluated setups and for each evaluated baseline (see Table~\ref{tab:real_robot_experiments} and Figure~\ref{fig:real_world_robot_learning_tasks}); the training hyperparameters are shown in Table~\ref{table:widowx_policies_training_configuration}. To train each of the three tasks performed with the WidowX robot, i.e., \text{Door Opening}, \text{Pick-and-Place}, and \text{Toy-Microwave Cooking}, we collected human demonstrations by teleoperating the robot with a Virtual Reality (VR) controller using the setup introduced in ~\cite{ebert2021bridge}; the number of collected demonstrations for each of the tasks is reported in Table~\ref{tab:data_collection_experimental_details}.
\input{tables/widowx_policies_training_configuration}
The policy's observations are RGB images and robot joint states. Images are encoded by a pre-trained visual encoder and a randomly initialized, unfrozen feature neck (``compression layer''~\cite{yadav2023ovrl}). We use the same feature neck as we did for the previously discussed MuJoCo tasks in Section~\ref{sec:simulation_results}. The encoded vector is concatenated with the robot's joint states, which is fed into a 3-layer MLP with a hidden dimension of size 256. The policy outputs end-effector commands, consisting of the end-effector's delta positions (Cartesian coordinates), delta rotations (Euler angles), and the gripper's opening/closing command; such commands are tracked by the robot at a frequency of 5 Hz. In addition to the hyperparameters listed in Table~\ref{table:widowx_policies_training_configuration}, we vary the policy action prediction horizon depending of the difficulty of the task, i.e., at each step the policy predicts the next 10, 10 and 5 actions for Door Opening, Toy-Microwave Cooking, and Pick-and-Place, respectively.

In the following, we give more details about the WidowX tasks showcased in our work.

\paragraph{Door Opening.} In this task the robot has to open a fridge door in a toy-kitchen setup; we identify two stages to evaluate the task's success: \textit{Open} and \textit{Fully Open} (see Figure~\ref{fig:real_world_robot_learning_tasks}). We place the robot in front of the fridge and collect 63 demonstrations to train the behavioral cloning policy. We vary the height (z-axis) of the robot base between 40-46cm, and the position (x-axis, parallel to the toy kitchen) of the robot base; samples from the initial state distribution of the demonstrations are shown in Figure~\ref{fig:initial_state_distribution}. At inference time, we vary the height of the robot base among $\{40, 42, 43, 44, 46\}$ cm, and select between 5 randomly-picked positions along the x-axis (for all policies). For each robot position, we evaluate the policy twice, for a total of 50 runs.

\input{figures/initial_state_distribution}

\paragraph{Pick-and-Place.} In this task, the robot has to pick up a pink cup from a toy-sink and drop it into a drying rack located on the left of the sink. We collected 48 demonstrations to train the policy, where we varied the initial pose of the objects, as shown in Figure~\ref{fig:initial_state_distribution}. During evaluation, we vary the cup's starting position amongst a total of 10 positions, of which 8 positions are equally distributed about the perimeter, and 2 are in the center of the sink. We also roughly vary the direction of the cup handle towards the left or the right. In total, we evaluate this task for 20 runs. There are two key stages for which we measure the success rate: picking up the cup and successfully releasing it into the drying rack.

\paragraph{Toy-Microwave Cooking.} In this task, the robot has to pick up an object from within the pot on the stovetop, putting the object into a toy-microwave, and closing the microwave. In each test, we initialize the environment with the microwave door open. In this task, we collected 100 demonstrations across 10 different toy-food objects (10 demonstrations per object) with randomized object positions; examples from the initial state distributions of the collected demonstrations can be seen in Figure~\ref{fig:initial_state_distribution}. During evaluation, we test 40 runs on 10 seen, in-distribution objects (4 runs per object), and 10 runs on 5 unseen, out-of-distribution objects (2 each), for a total of 50 runs. Furthermore, we vary the position of the pot that holds the object. We point out that this task is characterized by a longer horizon with respect to the other two; in fact, it has three different steps: 1. picking up the object, 2. placing the object into the microwave, 3. and closing the microwave. In addition, the policy needs to generalize to the different poses of the objects on the stove-top and to out-of-distribution objects. The longer horizon and the variety of objects in this task make it particularly challenging, so using frozen visual encoders was not effective (0\% success rate). However, with fine-tuning, the policies performed reasonably well.

\input{figures/initial_state_distribution_spot}

\input{tables/spot_policy_parameters}
\subsubsection{Spot Experiments}
\paragraph{Spot Setup.}
For the Spot experiments, we train a diffusion policy~\cite{chi2023diffusion} conditioned on the encoded image.  The diffusion policy outputs the desired absolute positions and rotations of the end-effector and the gripper state.  Hyperparameters for the policies are shown in Table~\ref{table:spot_policies_training_configuration}.
\paragraph{Drawer Opening.}
In this task, the robot has to open the top drawer of a cabinet.  The policy receives color images from a single forward facing camera mounted on the body of the Spot.  A fiducial marker is used to enable the robot to walk to a random position and orientation for each trial. Samples from the distribution of random initial states are shown in Figure~\ref{fig:initial_state_distribution_spot}. The starting locations vary by $\pm5$ cm in x and y and $\pm0.2$ radians in orientation.  50 successful demonstrations were collected using a scripted policy and we evaluate each policy for 20 trials.  A trial is considered successful if the drawer is opened at least 10 cm.  After each trial, a scripted policy uses the fiducial marker to reset the environment and the robot moves to a new random location. 

\section{More Visualizations of Translating to Teacher Model Features}
\input{figures/more_decoding_visualizations}
We attach examples of decoding Theia-representations of 4 frames from a robot video into VFM outputs in Figure~\ref{fig:more_decoding_visualizations}. Note that Theia and VFMs are not trained on the robot images on which we run this evaluation.

\section{Per-Task CortexBench Results}
In Table~\ref{tab:per_task_results_mujoco} we report per-task scores of the models evaluated in Figure~\ref{fig:mujoco_results}, over the MuJoCo subset of tasks. In Table~\ref{tab:per_task_results_cortexbench}, we report per-task scores of all 14 tasks we evaluated in Cortexbench, corresponding to Table~\ref{tab:main_cortexbench_table}. Note that we perform the evaluation following the original Cortexbench~\cite{majumdar2024we} protocol, where there are a total of 75 runs per MuJoCo task (we increased it from 30 to 75 for DMC tasks), 100 runs in Reach Cube, and 4200 runs in ImageNav.
\input{tables/per_task_results}

\section{Analysis of Visual Representations}\label{sec:analysis_visual_representations}
\input{figures/feature_norm_entropy_full_appendix}
\input{figures/feature_norm_map_clipped}
\input{figures/feature_norm_map_all_other_models}
\input{figures/feature_norm_distribution}
\paragraph{Entropy of the Representation Norm Distribution.}
Given $N$ representations produced by encoding $N$ images per model, where each representation contains $P$ spatial tokens, we discretize the distribution of token norms over all $N \mkern-5mu  \times \mkern-5mu  P$ tokens by using a  histogram. We normalize the count of each bin in histogram by the total number of tokens to obtain the probabilities of each bin. We then calculate the Shannon entropy, given by $-\sum_i p_i\log(p_i)$. We find that the distilled models have higher entropy than the regular models, so we divide them into two distinct groups. Results are plotted as model performance vs entropy on the MuJoCo tasks. We attach the full version of plot presented on the left of Figure~\ref{fig:feature_quality} here in Figure~\ref{fig:feature_norm_entropy_full_appendix}, including two plots corresponding to each category of models and one plot for all models. 

At the top of Figure~\ref{fig:feature_norm_map_clipped}, we find that both CLIP and VC-1 have high-norm outlier tokens. To better visualize the values of normal tokens, we use the median of norm values to clip the values. Specifically, we clip the norm values to range $[0, 2*\text{median}]$ and visualize the clipped norm values on the bottom of Figure~\ref{fig:feature_norm_map_clipped}. We find that the high-norm tokens are still not task-relevant.

In Figure~\ref{fig:feature_norm_map_all_other_models}, we attach the feature norm map of all other models. Among those, we find that MVP~\cite{xiao2022masked}, which performs well on CortexBench, also produces features without outlier tokens. Feature norms from Depth-Anything~\cite{yang2024depth} and SAM~\cite{kirillov2023segment}, in contrast, have low diversity.

We investigated the distributions of the feature norms that could give more information. Plots in Figure~\ref{fig:feature_norm_distribution} we show the histogram of the feature norm entropy of 4 selected models: ViT-B, DINOv2-B, VC-1-B, and Theia-B (refered to as rvfm\_deit\_base\_4features in the legend).

We find that Theia has the most balanced distribution of the feature norm, while the others are long-tailed. ViT-B has two peaks in the histogram. VC-1 is extremely centered and heavily long-tailed. DINOv2 is relatively less long-tailed than ViT-B and VC-1, and has a relatively nicer distribution around the median value.

\input{figures/feature_pcaauc_less}
\input{figures/feature_cossim_pcaauc_performance}
\paragraph{PCA Explained-Variance Ratio of Representations.} Similar to the entropy analysis, given $N \mkern-5mu  \times \mkern-5mu  P$ spatial token representations, we apply PCA to them and extract the explained-variance ratio (EVR) of each latent dimension. We calculate and plot the cumulative sum of EVRs, as well as calculate the Area Under the Curve (AUC) of cumulative sum of the EVR. When comparing Theia-B with ViT-B, DINOv2-B and VC-1-B (Figure~\ref{fig:feature_pca_evr_less}), we find that Theia-B has the lowest AUC and the best MuJoCo performance, while VC-1 has the highest AUC and the worse MuJoCo performance amongst these 4 models. The higher AUC is caused by one or few principle components that have very high EVRs, indicating that these components are capturing the majority of the variance of the feature representations. This means that less information is encoded within such representations. In contrast, the Theia-representation has a low AUC which we believe is due to the rich information that has been encoded within the latent space. 

However, when extending the scope to encompass all the models we evaluated (Figure~\ref{fig:feature_cossim_pcaauc_performance} left), we find that the AUC of the EVR does not have a strong correlation with robot learning performance. 

\paragraph{Cosine Similarity of Representations.}
We also use cosine similarity to analyze the representations from different models by first calculating the mean of all representations and then computing cosine similarity between each representation and this mean representation. Results are shown on the right of Figure~\ref{fig:feature_cossim_pcaauc_performance}, which shows very weak correlation between cosine similarity and performance on CortexBench.

\section{Linear Probing on ImageNet}
In addition to robot learning, we evaluate the Theia-representation on vision tasks to show to how well such abilities are maintained after the distillation process. For example, to evaluate image classification performance we apply linear probing on the Theia-representation to classify images from ImageNet-1k~\cite{jia2009imagenet}. We use mean pooling of the Theia-representation (i.e. spatial tokens) and the same training schedule as MAE~\cite{he2022masked}. Results are shown in Table~\ref{tab:imagenet_linear_probing}, where we find that Theia outperforms MAE~\cite{he2022masked} at the same model size, but is not comparable to SOTA results from models like DINOv2~\cite{oquab2023dinov2}.
\input{tables/imagenet_linear_probing}

\section{Probing 3D-awareness of Theia-representation}
We evaluate Theia-representation on 3D awareness to show the learned diverse vision knowledge in the representation. In particular, we adopt probe3D~\cite{elbanani2024probing}, an evaluation toolbox to investigate the 3D-awareness of pretrained models. This evaluation includes Monocular Depth Estimation, Surface Normal Estimation, and Multiview Correspondence.
We present the evaluation results in Tables~\ref{tab:depth_estimation}, \ref{tab:scannet_multiview_correspondence} and \ref{tab:surface_normal_estimation}, and we summarize them here:
\begin{itemize}
    \item Theia achieves the third best Depth Estimation performance among over 10 types of vision models (only worse than DeiT-III and DINOv2 models) (Table 2 in Appendix of probe3D~\cite{elbanani2024probing}, NYU dataset).
    \item Theia achieves comparable performance as the best performing model, DINOv2, on Multiview Correspondence on ScanNet (Table 4, Block3 in Appendix of probe3D~\cite{elbanani2024probing}).
    \item Theia has better Surface Normal Estimation performance than CLIP (Table 3 in Appendix of probe3D~\cite{elbanani2024probing}, NYU dataset).
\end{itemize}  
We note that the Theia model is trained only on ImageNet with 224x224 resolution images, while many baseline models in this benchmark are trained on billions of images and higher resolutions. Theia achieves better performance than models with similar training data and resolution.

\input{tables/probe3d}

\section{Generalization Evaluation}
We evaluate Theia-B and baseline models on a simulated benchmark, Factor-World~\cite{xie2023decomposing}, to test the generalization ability of the pre-trained visual representations. We use the Door-Open task from Factor-World benchmark environment, for which we collected 100 trajectories across 20 randomized environments. We trained the policies using the same policy head that we use in CortexBench MuJoCo tasks (Table~\ref{tab:policy_network}), and we evaluated the generalization ability when varying each of 6 factors of variations over 20 different runs controlled by one factor at a time. In Table~\ref{tab:generlization_factor_world}, we report the success rates of the baselines when varying each factor. These results show that Theia-B has the best generalization abilities compared to R3M, E-RADIO, and MVP. 
\input{tables/generalization_fw}
\end{document}

%% file: figures/performance_over_model_size.tex
\begin{figure}[thbp]
    \centering
    \vspace{-10pt}
    \includegraphics[trim={0 0 0 0},clip,width=\textwidth]{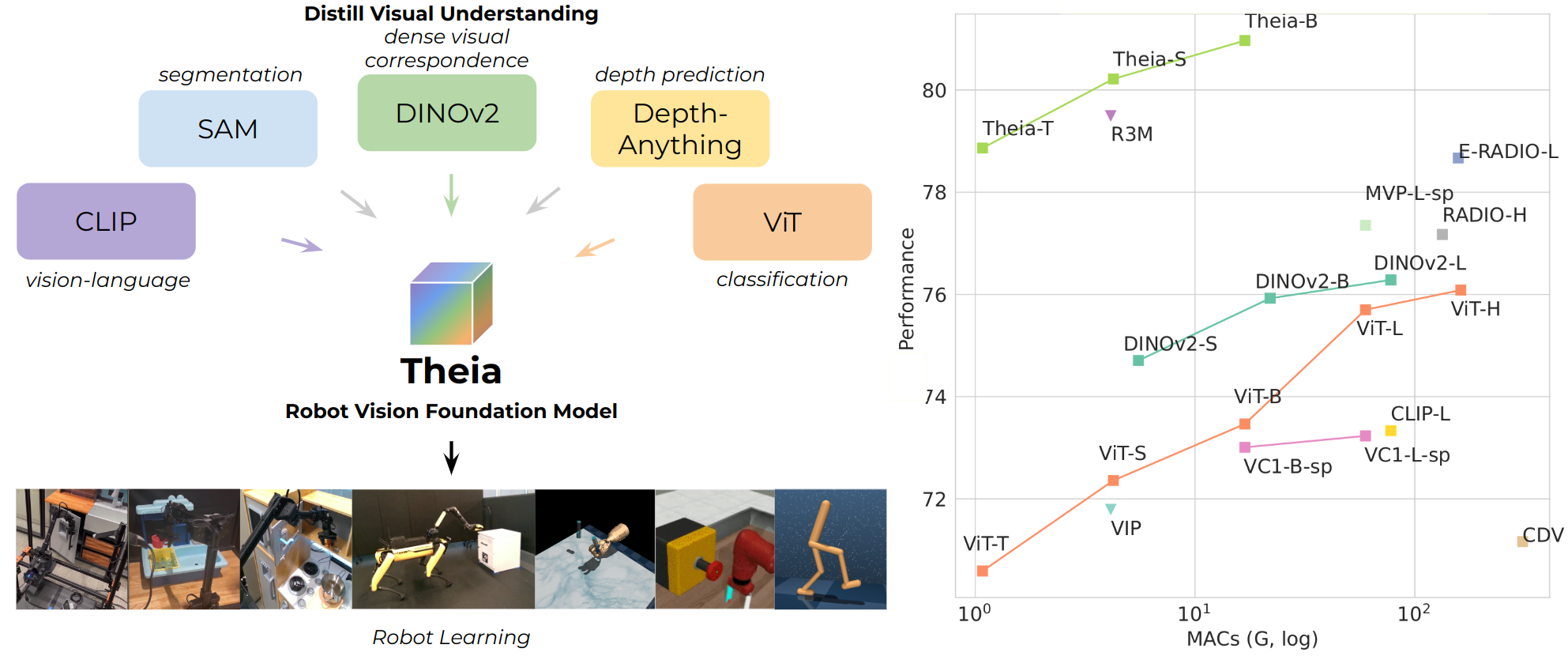}
    \vspace{-10pt}
    \caption{We introduce Theia, a model that distills multiple vision foundation models (VFMs) to provide better representations for robot learning (left). Theia achieves superior performance on robot learning tasks with less computation compared to standard VFMs and pre-trained models for robotics (right). Results shown are from the MuJoCo subset of tasks in CortexBench.}
    \label{fig:performance_over_model_size}
\end{figure}

%% file: figures/theia_main_figure.tex
\begin{figure}[t]
    \vspace{-30pt}
    \centering
    \includegraphics[width=\textwidth]{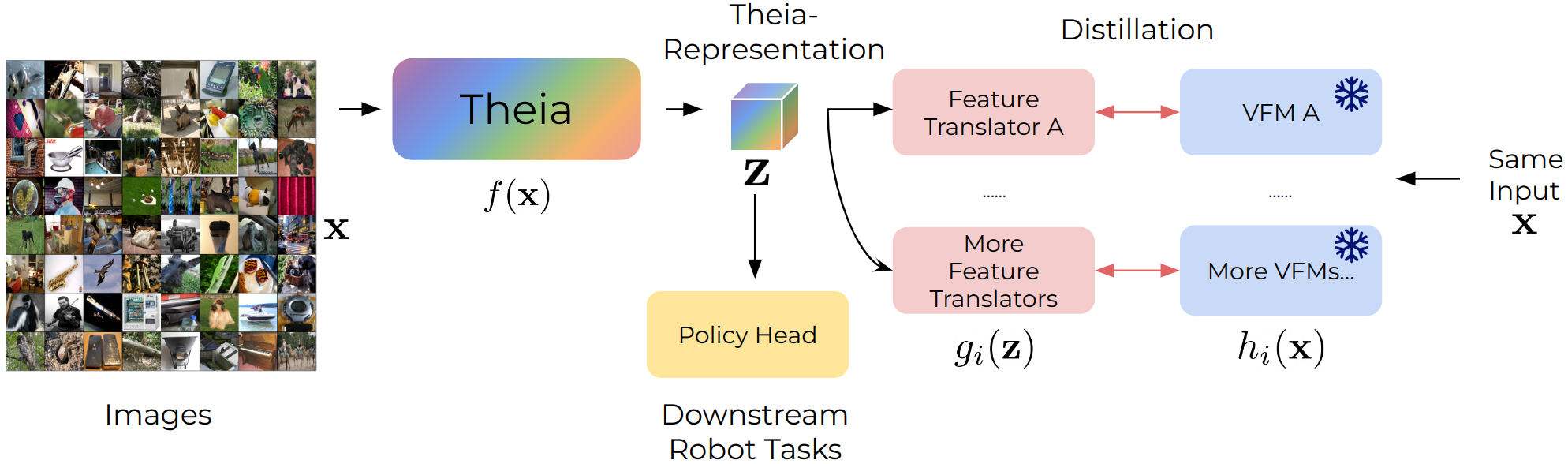}
    \caption{Theia distills multiple VFM features into one rich representation for robot learning. The feature translators $g_i(\mathbf{z})$ are supervised by the features from pretrained VFMs $h_i(\mathbf{x})$ during training time, then the distilled representation $\mathbf{z}$ is used as input to the policy head for robot learning tasks.}
    \label{fig:theia_main_figure}
    \vspace{-14pt}
\end{figure}

%% file: figures/robots_combined.tex
\begin{figure}[t]
   \centering
   \vspace{-20pt}
   \includegraphics[width=\textwidth]{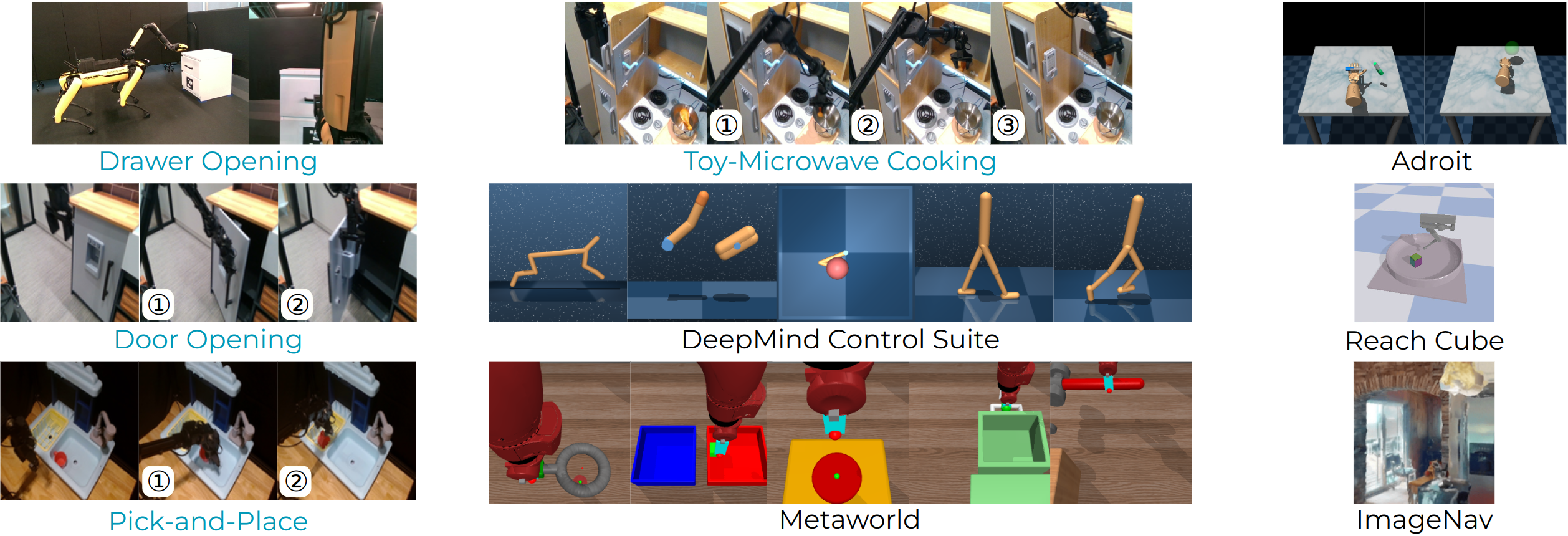}
   \vspace{-10pt}
   \caption{Simulation and real-world (labeled in blue) tasks used in this work. For simulated environments we show one image per task. For real-world tasks, we show images of key steps throughout the task labeled by numbers. A third-person view image shows the setup in Drawer Opening.} 
   \vspace{-17pt}
\label{fig:real_world_robot_learning_tasks}    
\end{figure}

%% file: figures/mujoco_results.tex
\setlength{\intextsep}{1pt}%
\setlength{\columnsep}{3pt}%
\begin{wrapfigure}[19]{r}{0.45\linewidth}
    \vspace{-7pt}
    \centering
    \captionsetup{width=0.43\textwidth}
    \includegraphics[width=\linewidth]{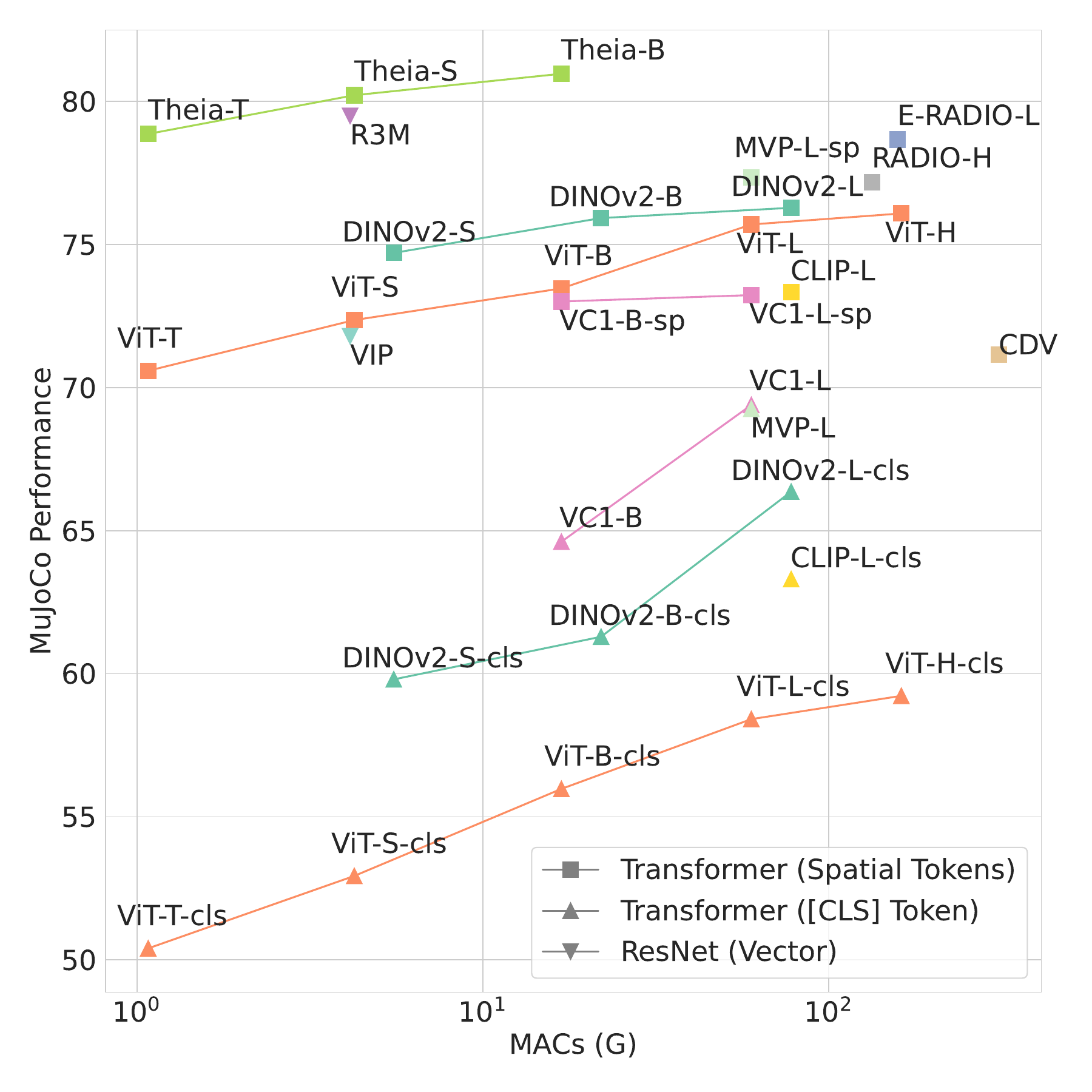}    
    \vspace{-20pt}
    \caption{Performance on MuJoCo tasks vs. inference computation. Theia achieves the best performance with much less compute (MACs (G): Multiply-Accumulate operations in billions, log-scale).}
    \label{fig:mujoco_results}
\end{wrapfigure}

%% file: tables/main_cortexbench_table.tex
\begin{table}[t]
    \vspace{-30pt}
    \centering
    \caption{Mean CortexBench score across 14 tasks (2 in Adroit, 5 in MetaWorld, 5 in DMC, 1 in Trifinger (Reach Cube), and 1 in Habitat (ImageNav)).}
    \resizebox{\textwidth}{!}{
    \begin{tabular}{ccccccc}
      \toprule
      Model  & \textbf{Theia-B (Ours)} & VC-1-L-sp~\cite{majumdar2024we} & MVP-L-sp~\cite{xiao2022masked} & R3M~\cite{nair2022r3m} & VIP~\cite{ma2022vip} & E-RADIO~\cite{ranzinger2023amradio} \\
      \midrule
      Score &  \textbf{79.79} $\pm$ \textbf{0.14} & 69.56 $\pm$ 0.80 & 77.42 $\pm$ 3.14 & 76.51 $\pm$ 0.79 & 71.18 $\pm$ 0.34 & 73.50 $\pm$ 1.69\\
      \bottomrule
    \end{tabular}
    }
    \label{tab:main_cortexbench_table}
\end{table}

%% file: tables/real_robot_experiments.tex
\begin{table}[t]
    \vspace{-13pt}
    \centering
    \caption{Real robot behavioral cloning results measured by success rate. Door Opening and Toy-Microwave Cooking are evaluated for 50 trials, and the others are evaluated for 20 trials. We report results for policies trained with either frozen (\Snowflake)  or fine-tuned (\protect\flame) visual encoders. For tasks having key stages (see Figure~\ref{fig:real_world_robot_learning_tasks}), we measure the success rates of achieving each stage separately.}
    \resizebox{\textwidth}{!}{
        \begin{tabular}{l c c c c c c c c c}
            \toprule
            \multirow{2}{*}{Model} & \multirow{2}{*}{\# Parameters (M)} & \multicolumn{2}{c}{\textbf{Door Opening \Snowflake}} & \multicolumn{2}{c}{\textbf{Pick-and-Place \Snowflake}} & \multicolumn{3}{c}{\textbf{Toy-Microwave Cooking} \flame} & \textbf{Drawer Opening} \\
            \cmidrule(l{2pt}r{2pt}){3-4} \cmidrule(l{2pt}r{2pt}){5-6} \cmidrule(l{2pt}r{2pt}){7-9} \cmidrule{10-10}
            & & \numcircledmod{1}~Open & \numcircledmod{2}~Fully Open & \numcircledmod{1}~Pick & \numcircledmod{2}~Place & \numcircledmod{1}~Pick & \numcircledmod{2}~Place & \numcircledmod{3}~Close the Door & \Snowflake~/ \flame \\
            \midrule
            \textbf{Theia-B (Ours)}             & 86 & \textbf{92}\% & \textbf{66}\% & \textbf{85}\% & \textbf{75}\% & \textbf{58}\% & 52\% & 40\% & \textbf{85}\% / \textbf{100}\% \\
            E-RADIO~\cite{ranzinger2023amradio} & 390  & 72\%          & 46\%          & 75\% & 55\%   & \textbf{58}\% & \textbf{54}\% & \textbf{42}\% & 15\% / 80\% \\
            MVP-L-sp~\cite{xiao2022masked}          & 303  &       32\% & 2\% & 55\% & 35\%          & 40\%          & 26\% & 18\%            & 30\% / 65\%\\
            VC-1-L-sp~\cite{majumdar2024we}     & 303  & 12\%          & 4\%           & 55\% & 45\%   & 14\% & 4\% & 4\%        & 0\% / 45\% \\
            R3M~\cite{nair2022r3m}              & 24 & 48\%          & 36\%          & 35\% & 10\%   & 36\% & 26\% & 18\%      & 0\% / 55\% \\
            DINOv2-L~\cite{oquab2023dinov2}     & 303  & 46\%          & 12\%          & 10\% & 0\%    & 20\% & 10\% & 2\%        & 35\% / 20\% \\
            ViT-H~\cite{dosovitskiy2021an}      & 632  & 18\%          & 4\%           & 15\% & 0\%    & 52\% & 44\% & 40\%      & 45\% / 85\% \\
            \bottomrule
        \end{tabular}
    }
    \label{tab:real_robot_experiments}
    \vspace{-15pt}
\end{table}

%% file: figures/combination_of_teacher_models.tex
\begin{figure}
    \centering
    \vspace{-30pt}
    \includegraphics[width=\textwidth]{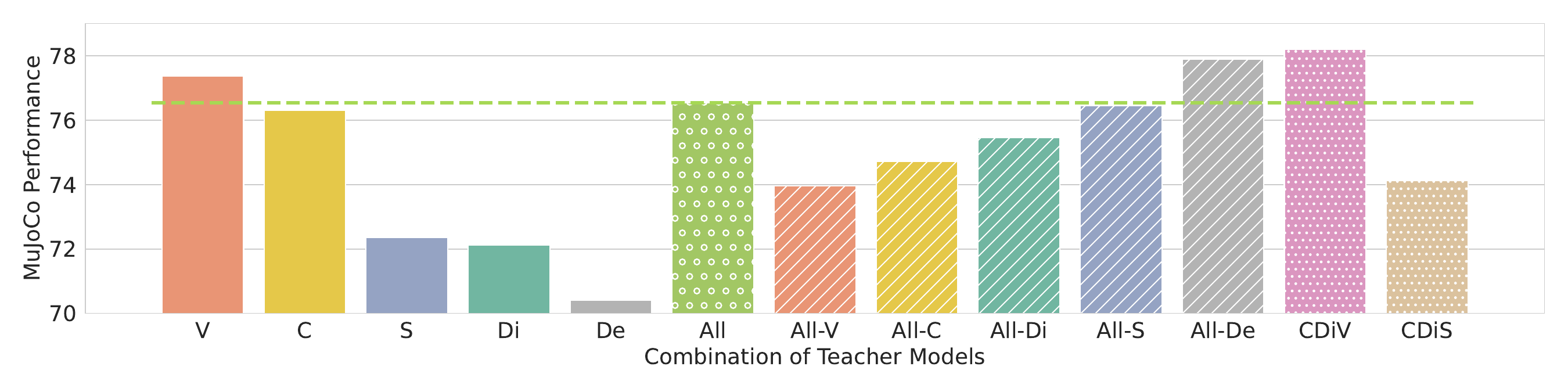}
    \vspace{-20pt}
    \caption{MuJoCo subset performance with respect to different combinations of teacher models to train Theia-T. Abbreviations of teacher models: \textbf{V}=ViT-H, \textbf{C}=CLIP-L, \textbf{S}=SAM-H, \textbf{Di}=DINOv2-L, \textbf{De}=Depth-Anything-L, \textbf{All}=all of five models (\textbf{CDeDiSV}), and \textbf{All}$-$\textbf{X}=taking \textbf{X} out of \textbf{All}.}
    \label{fig:combination_of_teacher_models}
    \vspace{-14pt}
\end{figure}

%% file: tables/ablation_studies.tex
\begin{table}[t]
    \centering
    \caption{Ablation studies on model design. All experiments are based on Theia-T unless otherwise labeled. Scores are the average performance from the MuJoCo subset of CortexBench.}
    \vspace{-6pt}
    \begin{subtable}[h]{0.20\textwidth}
        \centering
        \vspace{-8pt}
        \caption{Distillation losses}
        \vspace{-5pt}
        \setlength{\tabcolsep}{3pt}
        \resizebox{\textwidth}{!}{
            \begin{tabular}{cc}
            \toprule
            MSE &  Cos + L1 \\
            \midrule
            78.2 $\pm$ 1.0 & \textbf{78.9} $\pm$ 0.8 \\
            \bottomrule
            \end{tabular}
        }
        \label{tab:distillation_losses}
    \end{subtable}
    \begin{subtable}[h]{0.41\textwidth}
        \centering
        \vspace{-8pt}
        \caption{Number of register tokens (Theia-B)}
        \vspace{-5pt}
        \setlength{\tabcolsep}{3pt}
        \resizebox{0.95\textwidth}{!}{
            \begin{tabular}{cccc}
            \toprule
              0 & 1 & 4 & 8 \\
              \midrule
              79.2 $\pm$ 1.2 & \textbf{80.3} $\pm$ 0.7 & 79.4 $\pm$ 0.8 & 79.6 $\pm$ 0.9 \\
            \bottomrule
            \end{tabular}
        }
        \label{tab:num_register_tokens}
    \end{subtable}
    \begin{subtable}[h]{0.37\textwidth}
        \centering
        \caption{Tokens to Distill. SP=Spatial Tokens}
        \vspace{-4pt}
        \setlength{\tabcolsep}{4pt}
        \resizebox{\textwidth}{!}{
            \begin{tabular}{lcc}
            \toprule
             & Distill SP  & Distill \texttt{[CLS]} + SP \\
             \midrule
             Use \texttt{[CLS]} & --- & 50.0 $\pm$ 0.4 \\
             Use SP & \textbf{78.9} $\pm$ 0.8 & 77.3 $\pm$ 2.4 \\
            \bottomrule
            \end{tabular}
        }
        \label{tab:tokens_to_distill}
    \end{subtable}
    \label{tab:ablation_studies}
    \vspace{-15pt}
\end{table}

%% file: figures/decoding_visualization.tex

\begin{wrapfigure}{r}{0.5\textwidth}
  \centering
  \vspace{-10pt}
    \captionsetup{width=0.48\textwidth}
    \includegraphics[width=0.48\textwidth]{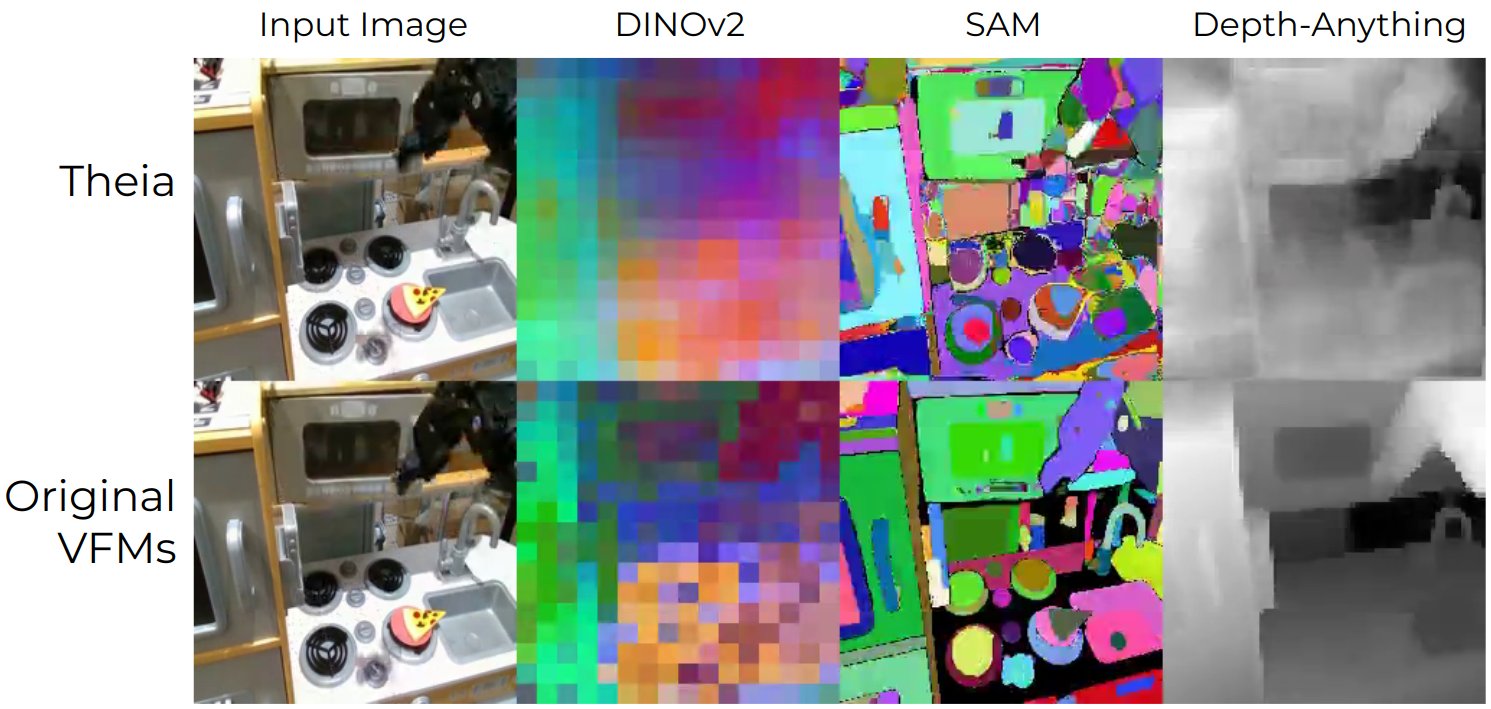}
    \vspace{-3pt}
    \caption{Visualization of VFM outputs from Theia predicted representations (top) and original VFM representations (bottom).}
    \label{fig:decoding_visualization}
\end{wrapfigure}

%% file: figures/feature_quality.tex
\begin{figure}[t]
    \vspace{-25pt}
    \centering
    \includegraphics[width=\textwidth]{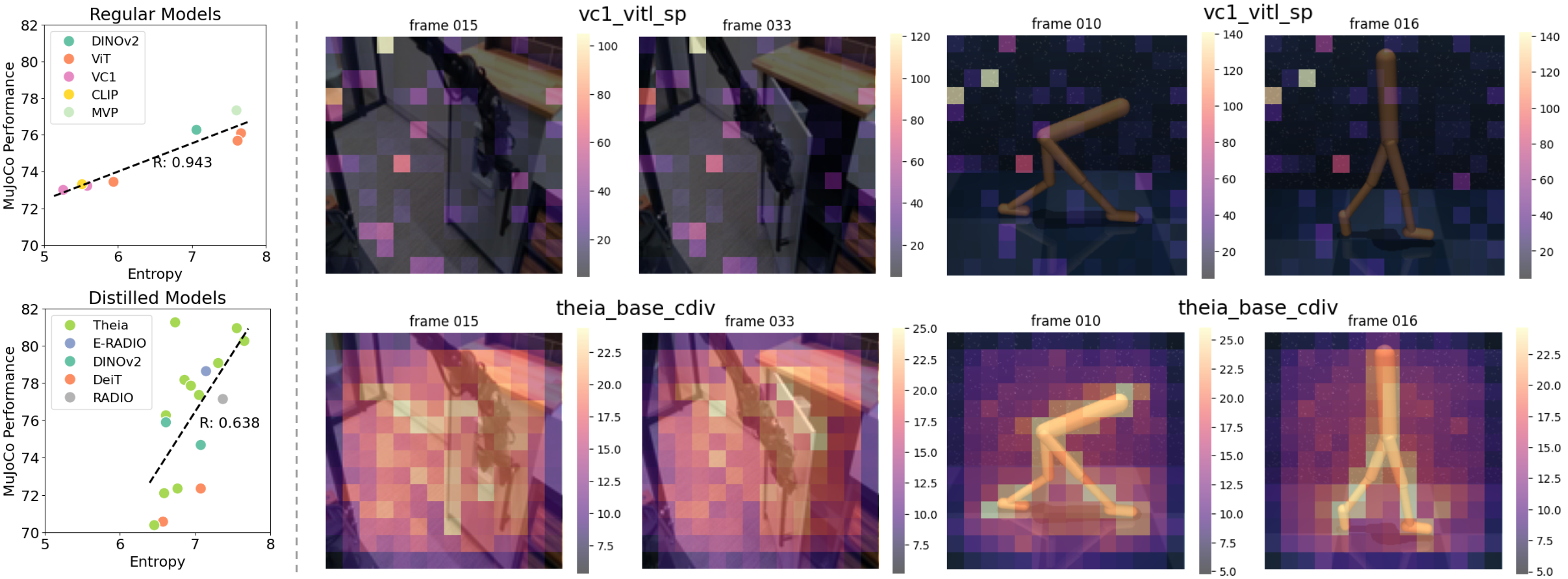}
    \caption{Left: Correlations between feature norm distribution entropy and robot learning performance; Right: Visualizations of spatial token feature norms from VC-1-L-sp and Theia-B.}
    \vspace{-13pt}
    \label{fig:feature_quality}
\end{figure}

%% file: tables/feature_translators.tex
\begin{table}[h]
    \centering
    \caption{Feature Translator configurations}
    \resizebox{\linewidth}{!}{
        \begin{tabular}{l}
        \toprule
        Student $d_s\times 14\times 14$ (Theia backbone=ViT-T, -S, -B) to Teacher $d_t\times 16\times 16$ (CLIP, DINOv2, ViT) \\
        \midrule
        ConvTranspose2d($d_s$, $d_s$, kernel\_size=3, stride=1, output\_padding=0)\\
        LayerNorm\\
        Conv2d($d_s$, $d_s$, kernel\_size=3, padding=1)\\
        ReLU+LayerNorm\\
        Conv2d($d_s$, $d_s$, kernel\_size=3, padding=1)\\
        ReLU+LayerNorm\\
        Flatten and Linear($d_s$, $d_t$)\\
        \toprule
        \\
        \toprule
        Student $d_s\times 14\times 14$ to Teacher $d_t\times 64\times 64$ (SAM and Depth-Anything) \\\midrule
        ConvTranspose2d($d_s$, $d_s$, kernel\_size=3, stride=2, padding=1\\
        LayerNorm\\
        ConvTranspose2d($d_s$, $d_s$, kernel\_size=3, stride=2, output\_padding=1)\\
        ReLU+LayerNorm\\
        Conv2d($d_s$, $d_s$, kernel\_size=3, padding=1)\\
        ReLU+LayerNorm\\
        Flatten and Linear($d_s$, $d_t$)\\
        \bottomrule
        \end{tabular}
    }
    \label{tab:feature_translators}
\end{table}

%% file: tables/training_configuration.tex
\begin{table}
    \centering
    \caption{Theia Training Configuration}
    \begin{tabular}{lc}
    \toprule
    Hyperparameters   & \\
    \midrule
    \# GPUs           &   8 \\
    Batch size        &   16 / GPU (128 effective) \\
    Learning rate (LR)&   5e-4 \\
    LR Schedule       &   Constant \\
    Weight decay      &   0.01 \\
    Optimizer         &   AdamW \\
    Betas             &   [0.9, 0.999] \\
    Epochs            &   50 \\
    Warm-up epochs    &   5 \\
    Warm-up LR schedule & Linear (1e-2*LR) \\
    Gradient clipping &   None \\
    Image augmentation &  None  \\
    Total GPU hours & 152 \\
    \bottomrule
    \end{tabular}
    \label{tab:training_configuration}
\end{table}

%% file: tables/ablation_studies_appendix.tex
\begin{table}[t]
    \centering
    \caption{More ablation studies on model design. All experiments are based on Theia-T. Scores are the average performance from the MuJoCo subset of CortexBench.}
    \vspace{-6pt}
    \begin{subtable}[h]{0.48\textwidth}
        \centering
        \caption{Feature Translator Architecture}
        \setlength{\tabcolsep}{3pt}
            \begin{tabular}{cc}
            \toprule
            CNNs &  Linear \\
            \midrule
            \textbf{78.9} $\pm$ 0.8 & 41.7 $\pm$ 1.6  \\
            \bottomrule
            \end{tabular}
        \label{tab:ablation_feature_translator}
    \end{subtable}
    \begin{subtable}[h]{0.48\textwidth}
        \centering
        \caption{Training from Scratch vs Pre-trained Backbone}
        \setlength{\tabcolsep}{3pt}
            \begin{tabular}{cc}
            \toprule
              Training from Scratch & Pre-trained Backbone \\
              \midrule
              78.9 $\pm$ 0.8 & \textbf{80.8} $\pm$ 1.5 \\
            \bottomrule
            \end{tabular}
        \label{tab:ablation_pretrained_backbone}
    \end{subtable}
    \label{tab:ablation_studies_appendix}

\end{table}

%% file: tables/detail_baseline_data_objective_comparison.tex
\begin{table}[th]
    \centering
    \caption{Comparison of model architectures, training datasets, total number of images, objectives, and training duration (epochs or GPU hours) across the models used in this paper. We use the numbers reported in their original papers and - stands for we could not find such information.}
    \resizebox{\textwidth}{!}{
    \begin{tabular}{cccccc}
    \toprule
       Model  & Architecture &  Dataset(s) & Total \# Images & Objective & Training Duration\\
    \midrule
       Theia & ViT & ImageNet-1k~\cite{jia2009imagenet} & 1.2M & Distillation & 50 epochs / 152 GPU hours on H100s \\
       RADIO / E-RADIO~\cite{ranzinger2023amradio} & ViT/Self-designed & DataComp-1B~\cite{gadre2024datacomp} & 1.4B & Distillation & - \\
       \cmidrule{1-6}
       VC-1~\cite{majumdar2024we} & ViT/MAE~\cite{he2022masked} & ImageNet-1k~\cite{jia2009imagenet}+V~\cite{grauman2022ego4d,goyal2017ssv2,damen2018epickitchens,shan2020100doh}+N & 5.6M & MAE~\cite{he2022masked} & 182 epochs / over 10,000 GPU hours \\
       MVP~\cite{xiao2022masked} & ViT/MAE~\cite{he2022masked} & ImageNet-1k~\cite{jia2009imagenet}+Video~\cite{damen2018epickitchens,shan2020100doh,goyal2017ssv2} & 1.9M  & MAE~\cite{he2022masked} & 1600 epochs \\
       \cmidrule{1-6}
       R3M~\cite{nair2022r3m} & ResNet~\cite{he2016deep} & Ego4D~\cite{grauman2022ego4d} & - & Time Contrastive~\cite{sermanet2018time}+Vision-Language Alignment & 1.5M steps \\
       VIP~\cite{ma2022vip} & ResNet~\cite{he2016deep} & Ego4D~\cite{grauman2022ego4d} & 4.3M & VIP~\cite{ma2022vip} & - \\
       \cmidrule{1-6}
       DINOv2~\cite{oquab2023dinov2} & ViT & LVD-142M~\cite{oquab2023dinov2} & 142M & Self-distillation & 22,016 GPU hours for DINOv2-g\\
       CLIP~\cite{radford2021learning} (Vision Encoder) & ViT & - & 400M & image-text contrastive learning~\cite{radford2021learning} & 73,728 GPU hours for CLIP ViT-L/14 on V100s\\
       ViT~\cite{dosovitskiy2021an} & ViT & ImageNet-21k~\cite{jia2009imagenet} / JFT-300M & 14M / 300M & Classification & 90 epochs / 7 epochs \\
       DeiT~\cite{touvron2021training} & ViT & ImageNet-1k & 1.2M & Classification+Distillation & 300 epochs / 288 GPU hours on V100s \\
       \bottomrule
    \end{tabular}
    }
    \label{tab:detail_baseline_data_objective_comparison}
\end{table}

%% file: tables/policy_network.tex
\begin{table}[th]
    \centering
    \caption{Policy Networks for MuJoCo Tasks}
    \begin{tabular}{l}
    \toprule
    Spatial Representation dimension $d\times H \times W$  \\
    \midrule
    Conv2d($d$, 256, kernel\_size=4, stride=2, padding=1) \\
    ReLU \\
    Conv2d(256, 256, kernel\_size=3, stride=2) \\
    ReLU \\
    Conv2d(256, 256, kernel\_size=3, stride=1) \\
    Flatten and Linear(256, 256) \\
    Linear(256, 256) \\
    Linear(256, action dimension) \\
    \midrule
    Vector Representation dimension $d$ \\
    \midrule
    Linear($d$, 256) \\
    Linear(256, 256) \\
    Linear(256, action dimension) \\
    \bottomrule
    \end{tabular}
    \label{tab:policy_network}
\end{table}

%% file: tables/experimental_details_table.tex
\begin{table}[t]
    \vspace{-20pt}
    \centering
    \caption{Number of demonstrations and data collection method for the real-world experiments.}
    \resizebox{\textwidth}{!}{
    \begin{tabular}{cccccc}
      \toprule
      Task  & \textbf{Door-Opening} & \textbf{Pick-and-Place} & \textbf{Toy-Microwave Cooking} & \textbf{Drawer Opening}\\
      \midrule
      \# Demos &  48 & 63 & 101 & 50 \\
      Method &  Teleoperation & Teleoperation & Teleoperation & Scripted Policy \\
      \bottomrule
    \end{tabular}
    }
    \label{tab:data_collection_experimental_details}
\end{table}

%% file: tables/widowx_policies_training_configuration.tex
\begin{table}
\centering
    \caption{WidowX Policy Training Configuration}
    \begin{tabular}{lc}
    \toprule
    Hyperparameters   & \\
    \midrule
    Batch size        &   16 \\
    Learning rate     &   1e-4 \\
    Weight decay      &   0.01 \\
    Optimizer         &   AdamW \\
    Betas             &   [0.9, 0.999] \\
    Epochs            &   400 \\
    Loss              &   SmoothL1 \\
    \bottomrule
    \end{tabular}
\label{table:widowx_policies_training_configuration}
\end{table}

%% file: figures/initial_state_distribution.tex
\begin{figure}[t]
\centering
    \begin{subfigure}[b]{1.0\linewidth}
        \includegraphics[width=\linewidth]{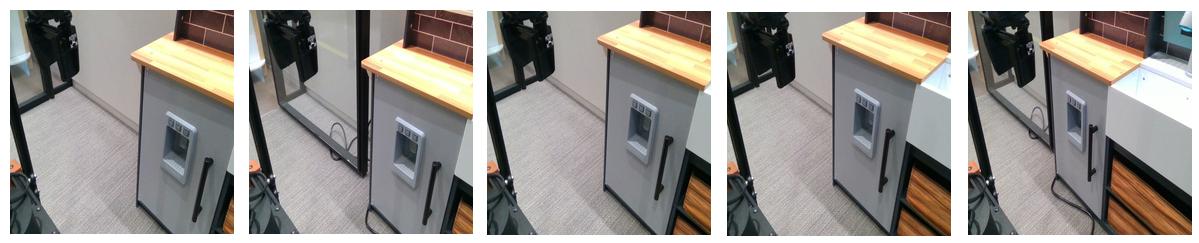}
    \end{subfigure}\vfill
    \begin{subfigure}[b]{1.0\linewidth}
        \includegraphics[width=\linewidth]{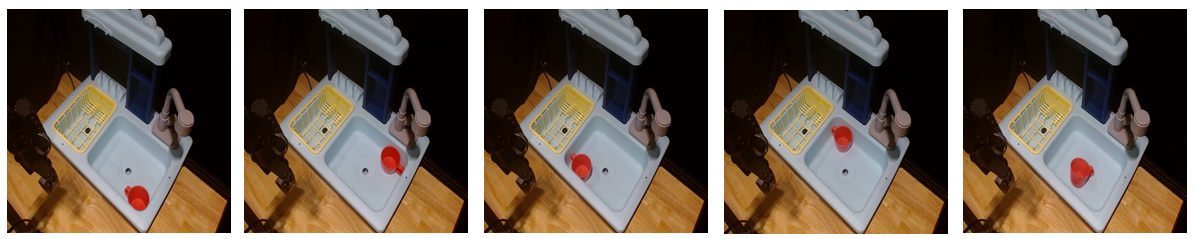}
    \end{subfigure}\vfill
    \begin{subfigure}[b]{1.0\linewidth}
        \includegraphics[width=\linewidth]{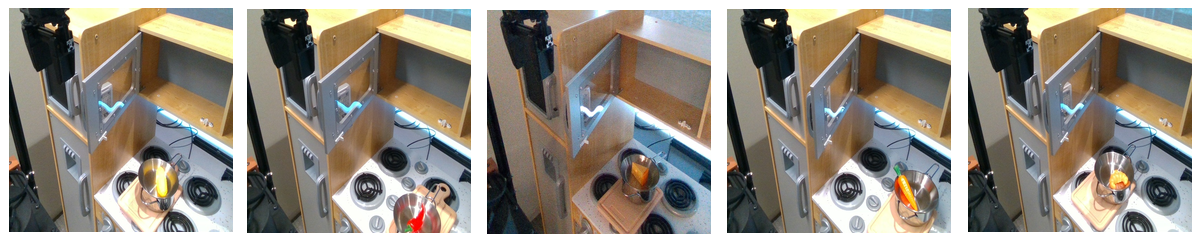}
    \end{subfigure}
    \caption{Samples from the initial state distribution of the collected demonstrations for \textit{fridge door-opening} (top), \textit{pick-and-place} (middle), and \textit{toy-microwave cooking} (bottom). For each of these tasks, the factors of variation are, respectively: the robot height and position with respect to the door, the initial pose of the cup, and the initial position of the objects and object types.}
    \label{fig:initial_state_distribution}
\end{figure}

%% file: figures/initial_state_distribution_spot.tex
\begin{figure}[t]
\centering
    \begin{subfigure}[b]{.24\linewidth}
        \includegraphics[width=\linewidth]{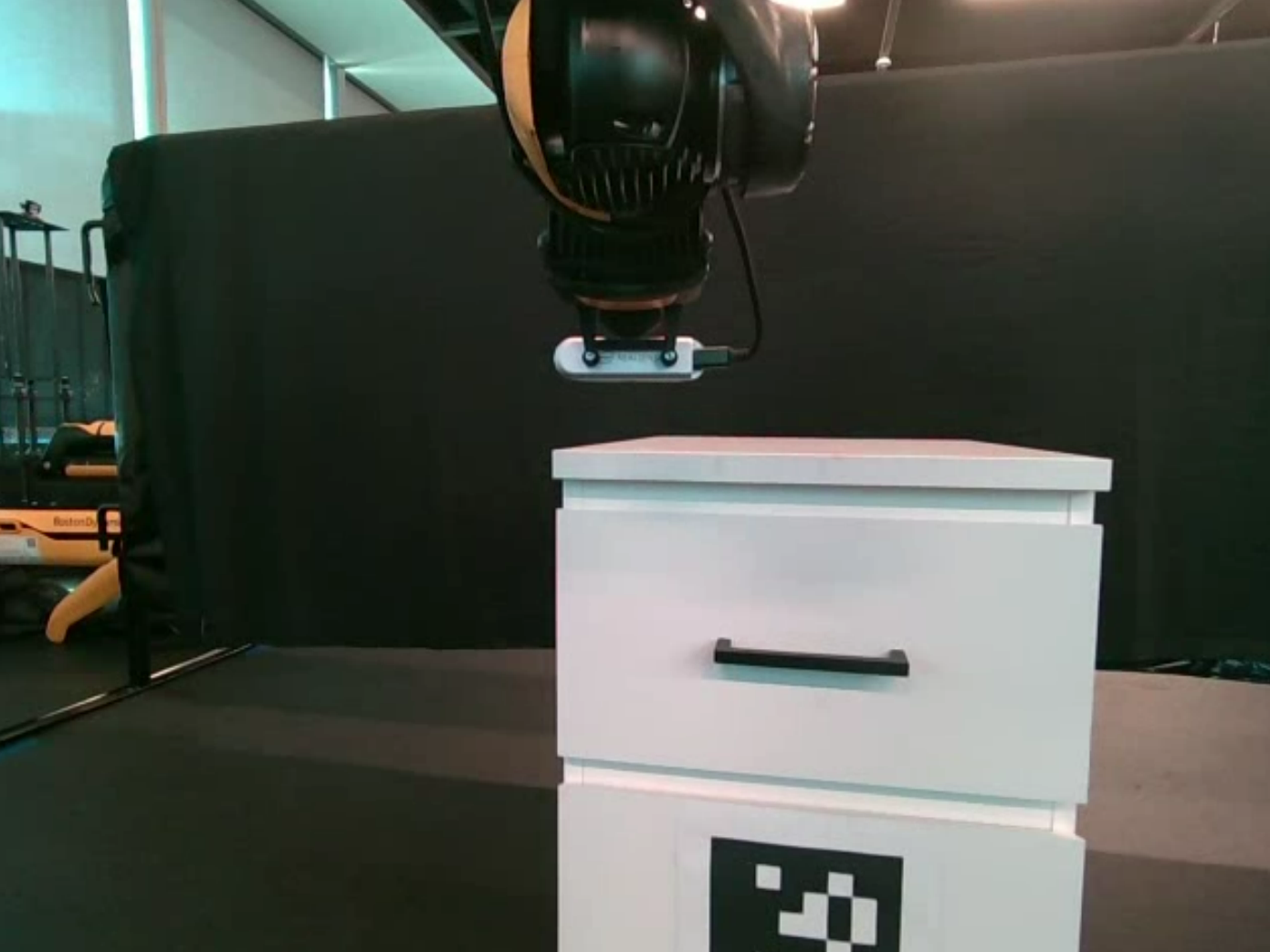}
    \end{subfigure}
    \begin{subfigure}[b]{.24\linewidth}
        \includegraphics[width=\linewidth]{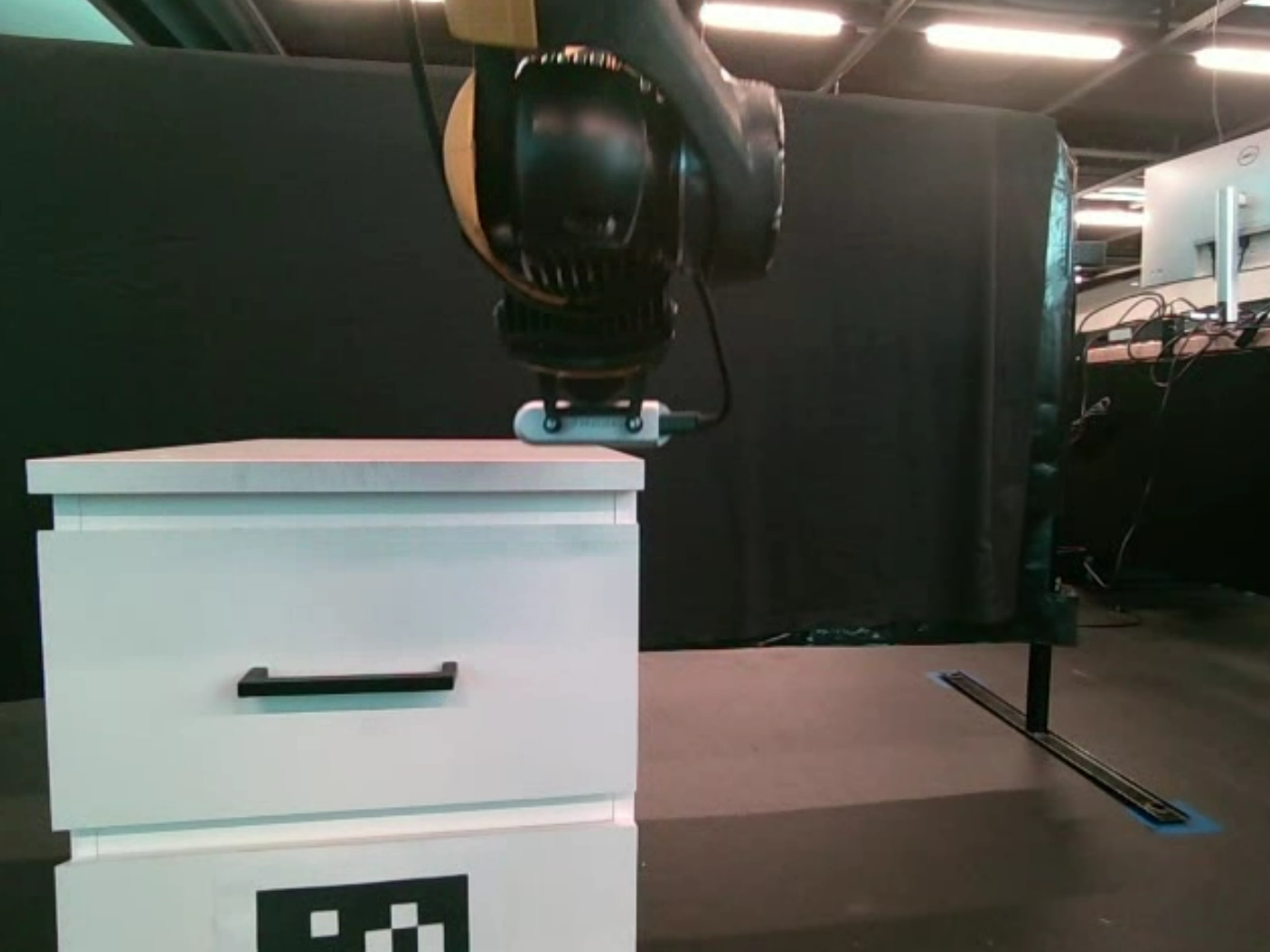}
    \end{subfigure}
    \begin{subfigure}[b]{.24\linewidth}
        \includegraphics[width=\linewidth]{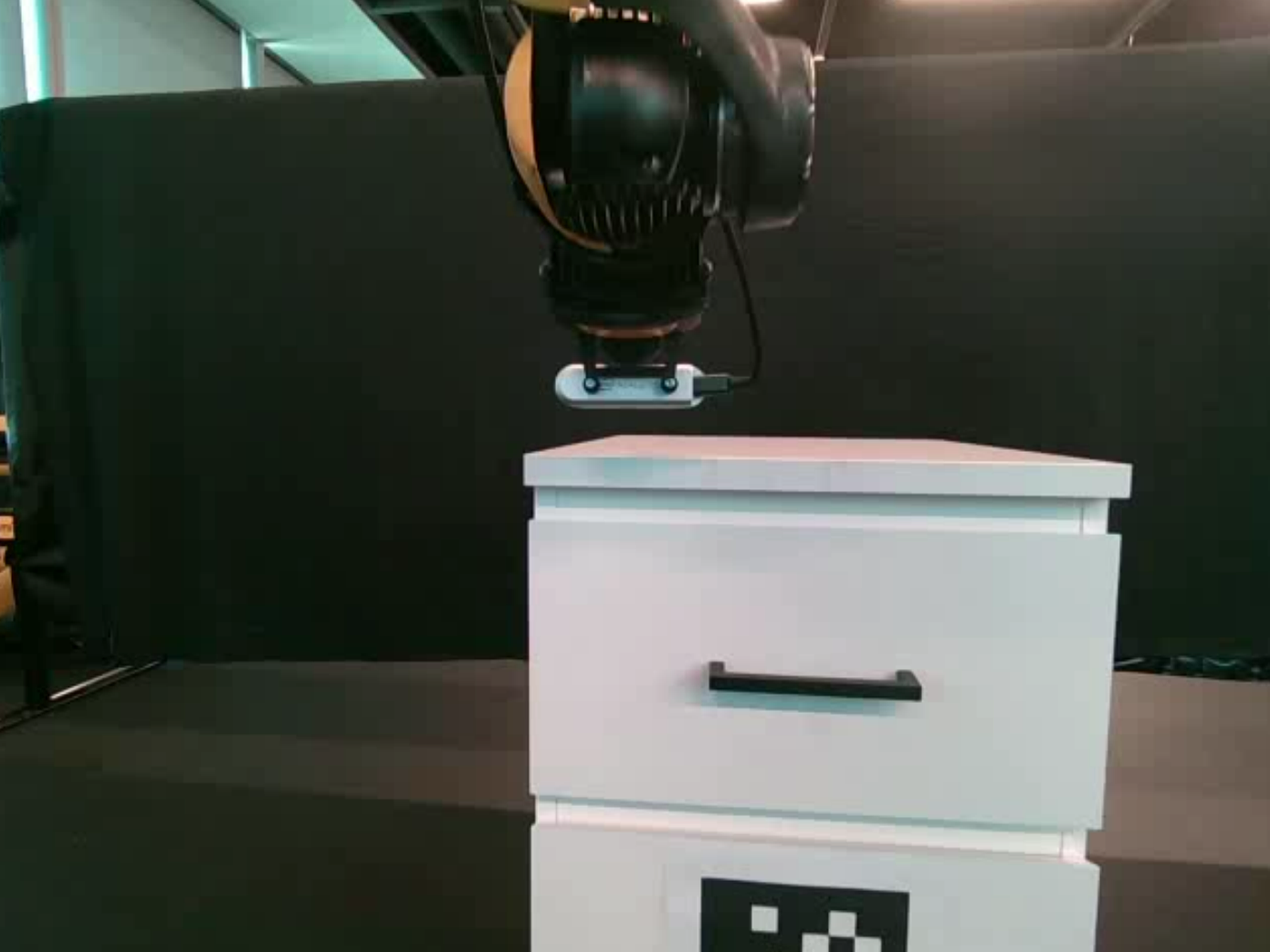}
    \end{subfigure}
    \begin{subfigure}[b]{.24\linewidth}
        \includegraphics[width=\linewidth]{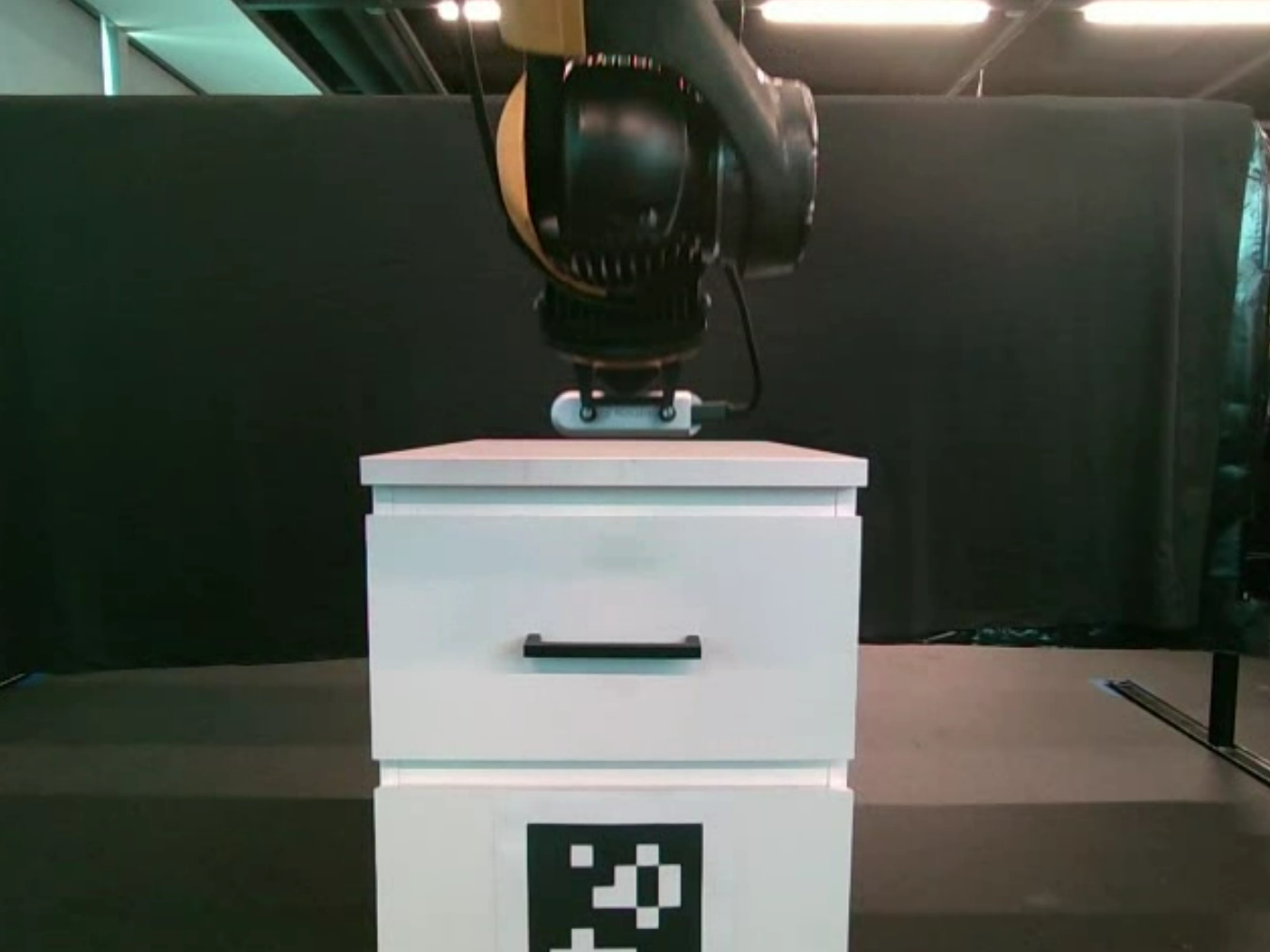}
    \end{subfigure}
    \caption{Samples from the initial state distribution of the collected demonstrations for the Spot drawer opening experiments. For this task, the factors of variation are the position and orientation of the robot with respect to the drawer.}
    \label{fig:initial_state_distribution_spot}
\end{figure}

%% file: tables/spot_policy_parameters.tex
\begin{table}
\centering
    \caption{Spot Policy Training Configuration}
    \begin{tabular}{lc}
    \toprule
    Hyperparameters   & \\
    \midrule
    Batch size        &   32 \\
    Learning rate     &   3e-4\\
    Weight decay      &   cosine \\
    Optimizer         &   Adam \\
    Betas             &   [0.9, 0.999] \\
    Training Iterations            &   3000 \\
    Loss              &   MSE \\
    Policy Horizon              &   4 \\
    \bottomrule
    \end{tabular}
\label{table:spot_policies_training_configuration}
\end{table}

%% file: figures/more_decoding_visualizations.tex
\begin{figure}[th]
    \centering
    \includegraphics[width=\textwidth]{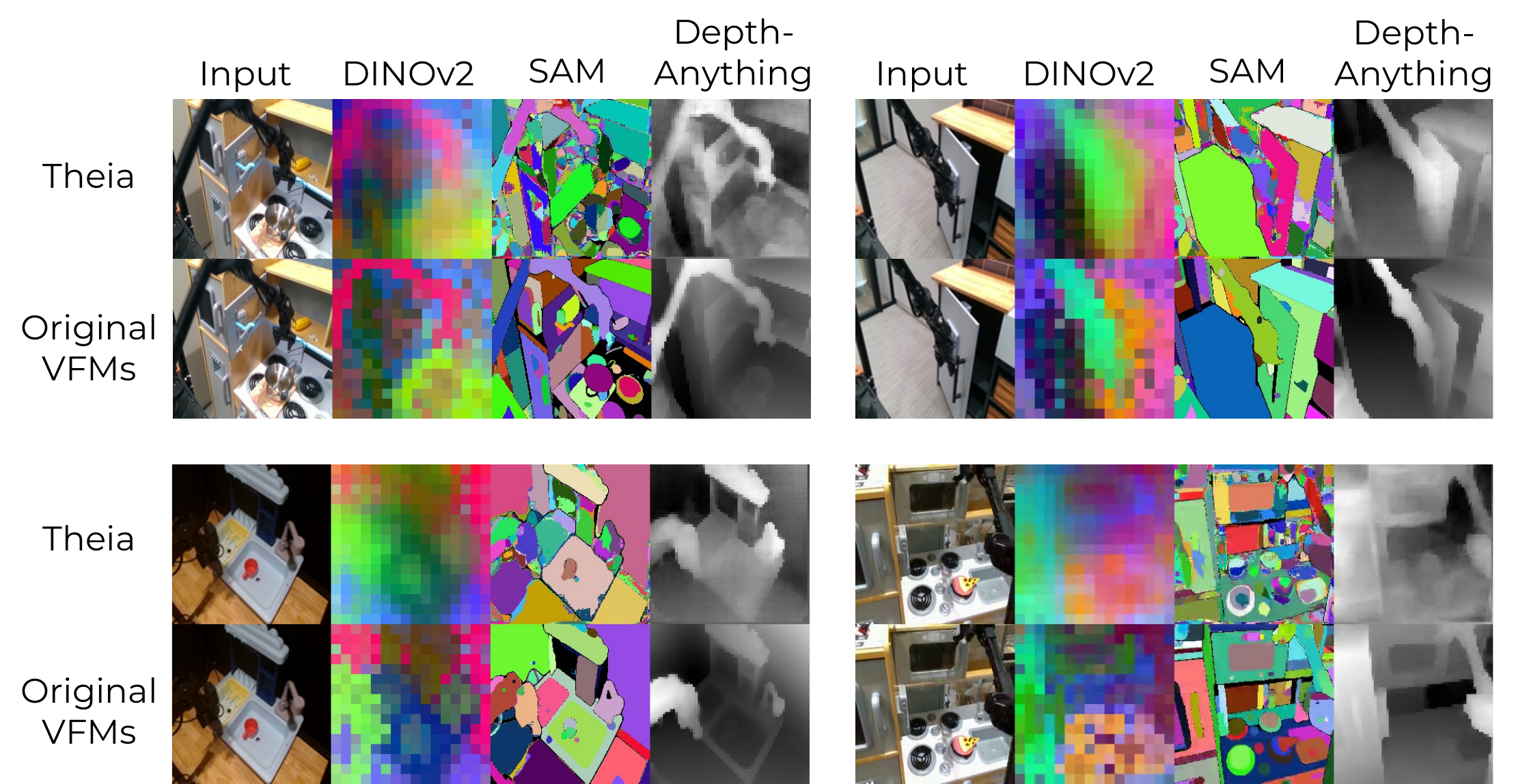}
    \caption{More examples of decoding Theia-representation to VFM outputs using feature translator and original VFM decoders. We select robot images from our experiment recordings. Theia and VFMs are not trained on these images.}
    \label{fig:more_decoding_visualizations}
\end{figure}

%% file: tables/per_task_results.tex
\begin{table}[ht]
    \centering
    \caption{Per-Task Results on the MuJoCo Subset.}
    \resizebox{\textwidth}{!}{
        \begin{tabular}{lcccccccccccc}
        \toprule
        Model & Assembly & Bin-picking & Button-press & Cheetah-run & Finger-spin & Reacher-easy & Walker-stand & Walker-walk & Drawer-open & Hammer & Pen & Relocate \\
        \toprule
        Theia-T & 90.00±8.49 & 74.00±8.49 & 80.00±0.00 & 63.78±1.65 & 69.76±0.97 & 79.18±3.87 & 90.59±1.60 & 81.06±1.92 & 100.00±0.00 & 98.00±2.83 & 74.00±2.83 & 46.00±2.83 \\
        Theia-S & 94.67±9.24 & 70.67±11.55 & 72.00±14.42 & 67.37±4.25 & 70.46±0.80 & 83.25±5.19 & 92.24±0.75 & 82.62±1.91 & 100.00±0.00 & 97.33±4.62 & 81.33±2.31 & 50.67±6.11 \\
        Theia-B & 93.33±8.33 & 76.00±4.00 & 82.67±6.11 & 67.67±1.92 & 70.84±1.37 & 83.23±7.05 & 92.55±3.67 & 81.33±3.11 & 100.00±0.00 & 98.67±2.31 & 78.67±2.31 & 46.67±2.31 \\
        DINOv2-L & 93.33±8.33 & 80.00±8.00 & 61.33±2.31 & 45.66±5.69 & 70.95±0.25 & 74.24±16.02 & 92.84±4.55 & 83.70±1.34 & 100.00±0.00 & 100.00±0.00 & 77.33±4.62 & 36.00±0.00 \\
        DINOv2-B & 92.00±8.00 & 76.00±14.42 & 72.00±4.00 & 48.02±3.71 & 70.77±0.59 & 75.84±2.63 & 92.64±1.81 & 83.82±2.26 & 100.00±0.00 & 98.67±2.31 & 68.00±4.00 & 33.33±2.31 \\
        DINOv2-S & 93.33±8.33 & 68.00±8.00 & 81.33±12.22 & 45.43±5.33 & 70.70±0.81 & 59.86±7.73 & 88.21±1.90 & 77.62±6.32 & 100.00±0.00 & 98.67±2.31 & 77.33±4.62 & 36.00±4.00 \\
        CLIP-L & 69.33±4.62 & 76.00±4.00 & 64.00±8.00 & 33.63±1.21 & 69.97±2.02 & 89.42±3.92 & 95.12±0.89 & 75.87±5.33 & 100.00±0.00 & 96.00±4.00 & 73.33±8.33 & 37.33±6.11 \\
        ViT-H & 94.67±9.24 & 76.00±12.00 & 77.33±9.24 & 47.89±10.10 & 69.84±1.16 & 84.33±6.46 & 90.97±5.36 & 79.99±6.24 & 100.00±0.00 & 94.67±2.31 & 56.00±0.00 & 41.33±4.62 \\
        ViT-L & 96.00±5.66 & 54.00±25.46 & 81.60±3.58 & 50.32±5.24 & 69.54±0.80 & 84.49±3.26 & 89.43±2.30 & 77.43±1.80 & 100.00±0.00 & 96.00±3.27 & 68.00±3.27 & 41.60±8.29 \\
        ViT-B & 96.00±6.93 & 74.67±10.07 & 76.00±10.58 & 46.28±5.32 & 72.05±1.23 & 73.71±0.80 & 77.13±6.50 & 69.75±3.36 & 100.00±0.00 & 96.00±4.00 & 68.00±0.00 & 32.00±4.00 \\
        ViT-S & 93.33±8.33 & 77.33±2.31 & 61.33±6.11 & 42.99±1.13 & 70.71±0.68 & 70.46±2.59 & 88.34±2.63 & 67.83±4.00 & 100.00±0.00 & 90.67±9.24 & 68.00±4.00 & 37.33±4.62 \\
        ViT-T & 93.33±8.33 & 82.67±4.62 & 65.33±10.07 & 39.02±1.47 & 71.45±1.07 & 71.13±3.24 & 84.05±0.84 & 70.74±3.03 & 100.00±0.00 & 84.00±6.93 & 60.00±10.58 & 25.33±4.62 \\
        VC-1-L-sp & 85.33±8.33 & 66.67±12.22 & 56.00±8.00 & 66.88±6.66 & 71.19±0.67 & 70.67±8.36 & 93.43±6.08 & 83.28±2.40 & 100.00±0.00 & 93.33±2.31 & 68.00±0.00 & 24.00±8.00 \\
        CDV & 73.33±24.44 & 74.40±11.87 & 30.67±39.26 & 50.53±12.08 & 71.94±1.07 & 71.60±17.03 & 94.85±1.27 & 78.36±5.72 & 100.00±0.00 & 93.60±6.07 & 75.33±5.89 & 39.33±10.25 \\
        RADIO & 96.00±6.93 & 84.00±8.00 & 82.67±12.86 & 35.92±1.59 & 71.98±1.41 & 78.46±6.50 & 89.06±2.42 & 81.33±4.91 & 100.00±0.00 & 98.67±2.31 & 68.00±0.00 & 40.00±6.93 \\
        E-RADIO & 94.67±9.24 & 82.67±2.31 & 80.00±4.00 & 57.37±2.27 & 69.68±1.05 & 75.59±1.29 & 91.73±2.03 & 80.32±2.06 & 100.00±0.00 & 100.00±0.00 & 66.67±12.86 & 45.33±2.31 \\
        MVP-L-sp & 93.33±8.33 & 73.33±4.62 & 82.67±10.07 & 68.07±1.71 & 71.03±2.10 & 69.87±6.75 & 88.44±4.21 & 80.14±1.50 & 100.00±0.00 & 97.33±2.31 & 77.33±12.22 & 26.67±11.55 \\
        MVP-L & 94.67±9.24 & 82.67±9.24 & 89.33±8.33 & 34.62±5.83 & 68.63±2.02 & 67.95±3.18 & 74.50±1.65 & 48.04±1.37 & 100.00±0.00 & 88.00±6.93 & 62.67±6.11 & 20.00±4.00 \\
        R3M & 96.00±6.93 & 92.00±4.00 & 68.00±4.00 & 55.88±1.12 & 70.65±0.34 & 82.37±3.70 & 88.88±2.70 & 69.52±4.94 & 100.00±0.00 & 98.67±2.31 & 73.33±2.31 & 58.67±4.62 \\
        VIP & 93.33±8.33 & 70.67±8.33 & 76.00±4.00 & 45.10±4.02 & 69.02±0.67 & 68.08±3.45 & 78.50±2.49 & 63.52±1.40 & 98.67±2.31 & 96.00±4.00 & 73.33±6.11 & 29.33±10.07 \\
        \bottomrule
        \end{tabular}        
    }
\label{tab:per_task_results_mujoco}
\end{table}

\begin{table}[t]
    \centering
    \caption{Per-task results on CortexBench (excluding Move Cube, ObjectNav, and MobilePick due to reproducibility issues).}
    \resizebox{\textwidth}{!}{
        \begin{tabular}{lccccccccccccccc}
        \toprule
        Model & Assembly & Bin-picking & Button-press & Cheetah-run & Finger-spin & Reacher-easy & Walker-stand & Walker-walk & Drawer-open & Hammer & Pen & Relocate & Reach-cube & ImageNav \\
        \toprule
        Theia-B & 93.33±8.33 & 76.00±4.00 & 82.67±6.11 & 67.67±1.92 & 70.84±1.37 & 83.23±7.05 & 92.55±3.67 & 81.33±3.11 & 100.00±0.00 & 98.67±2.31 & 78.67±2.31 & 46.67±2.31 & 86.19±0.11 & 59.3±0.7  \\
        VC-1-L-sp & 85.33±8.33 & 66.67±12.22 & 56.00±8.00 & 66.88±6.66 & 71.19±0.67 & 70.67±8.36 & 93.43±6.08 & 83.28±2.40 & 100.00±0.00 & 93.33±2.31 & 68.00±0.00 & 24.00±8.00  & 84.79±0.63 & 70.3±0.7 \\
        E-RADIO & 94.67±9.24 & 82.67±2.31 & 80.00±4.00 & 57.37±2.27 & 69.68±1.05 & 75.59±1.29 & 91.73±2.03 & 80.32±2.06 & 100.00±0.00 & 100.00±0.00 & 66.67±12.86 & 45.33±2.31  & 87.81±0.12 & 53.0±0.7 \\
        MVP-L-sp & 93.33±8.33 & 73.33±4.62 & 82.67±10.07 & 68.07±1.71 & 71.03±2.10 & 69.87±6.75 & 88.44±4.21 & 80.14±1.50 & 100.00±0.00 & 97.33±2.31 & 77.33±12.22 & 26.67±11.55 & 87.54±0.2 & 68.1±0.7 \\
        R3M & 96.00±6.93 & 92.00±4.00 & 68.00±4.00 & 55.88±1.12 & 70.65±0.34 & 82.37±3.70 & 88.88±2.70 & 69.52±4.94 & 100.00±0.00 & 98.67±2.31 & 73.33±2.31 & 58.67±4.62 & 86.5 & 30.6±0.7\\
        VIP & 93.33±8.33 & 70.67±8.33 & 76.00±4.00 & 45.10±4.02 & 69.02±0.67 & 68.08±3.45 & 78.50±2.49 & 63.52±1.40 & 98.67±2.31 & 96.00±4.00 & 73.33±6.11 & 29.33±10.07  & 86.2 & 48.8±0.8\\
        \bottomrule
        \end{tabular}        
    }
\label{tab:per_task_results_cortexbench}
\end{table}

%% file: figures/feature_norm_entropy_full_appendix.tex
\begin{figure}[th]
    \centering
    \includegraphics[width=0.7\textwidth]{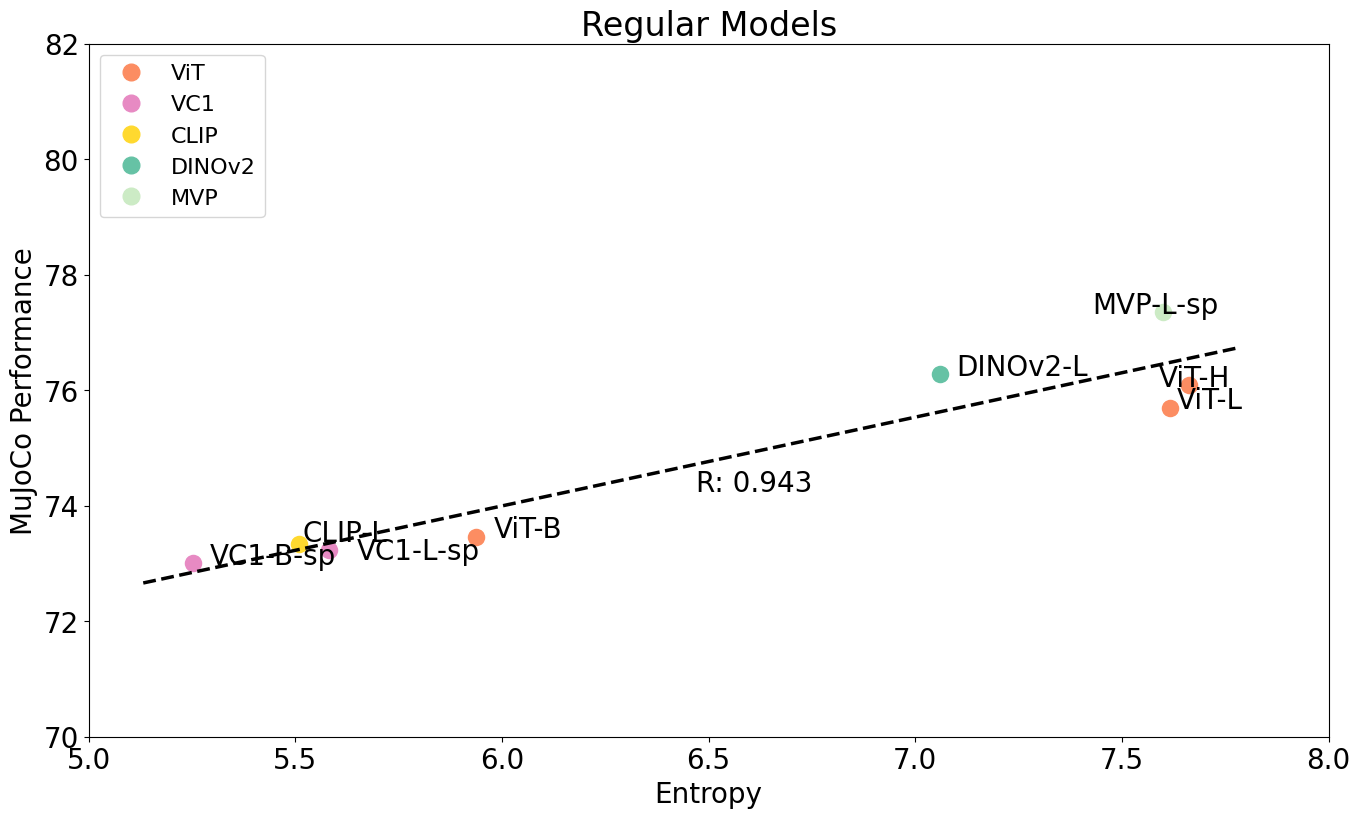}
    \includegraphics[width=0.7\textwidth]{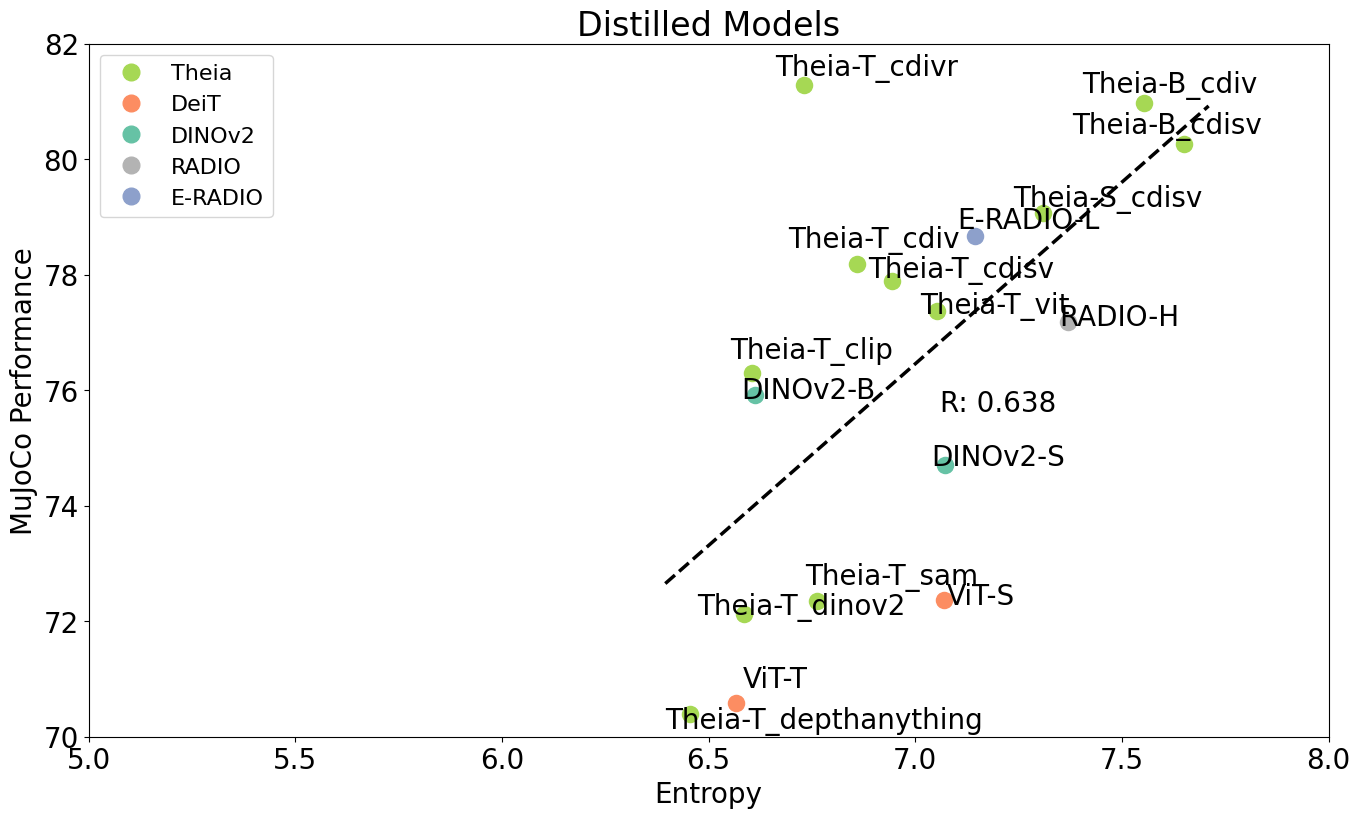}
    \includegraphics[width=0.7\textwidth]{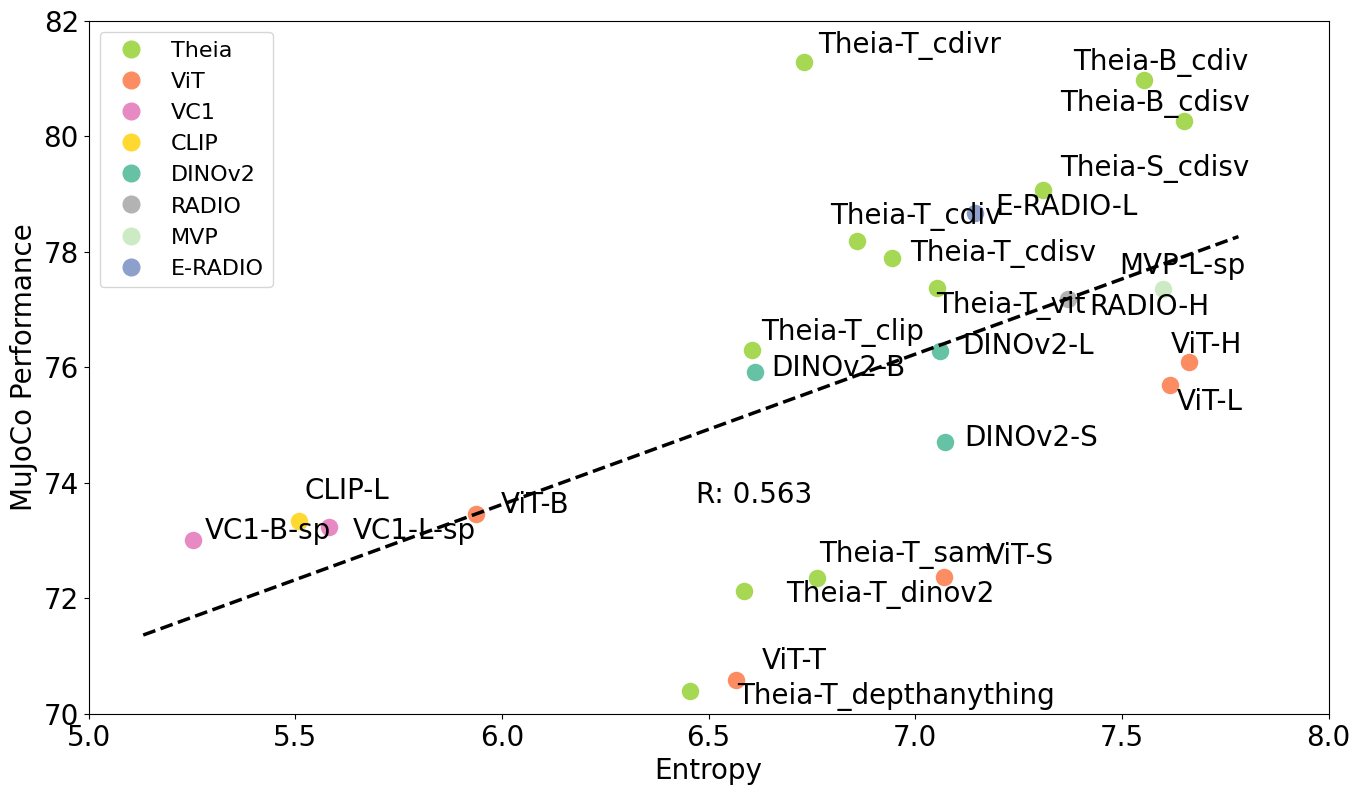}
    \caption{Full results of feature norm entropy.}
    \label{fig:feature_norm_entropy_full_appendix}
\end{figure}

%% file: figures/feature_norm_map_clipped.tex
\begin{figure}[th]
    \centering
    \includegraphics[width=\textwidth]{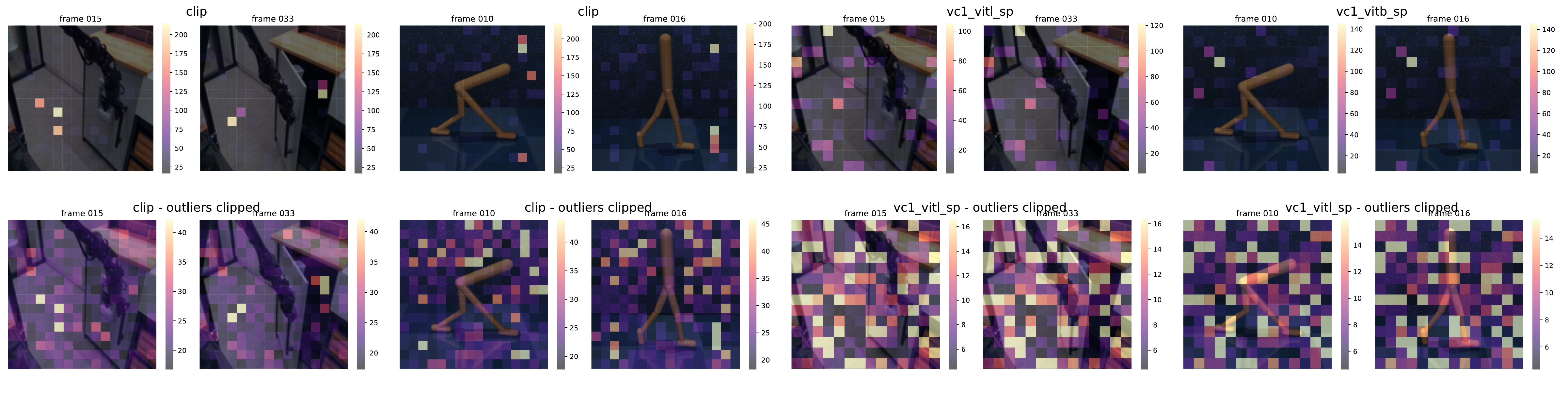}
    \caption{Feature norm map visualizations of CLIP and VC1, original (top) and with high-norm outlier values clipped (bottom)}
    \label{fig:feature_norm_map_clipped}
\end{figure}

%% file: figures/feature_norm_map_all_other_models.tex
\begin{figure}[th]
    \centering
    \includegraphics[width=\textwidth]{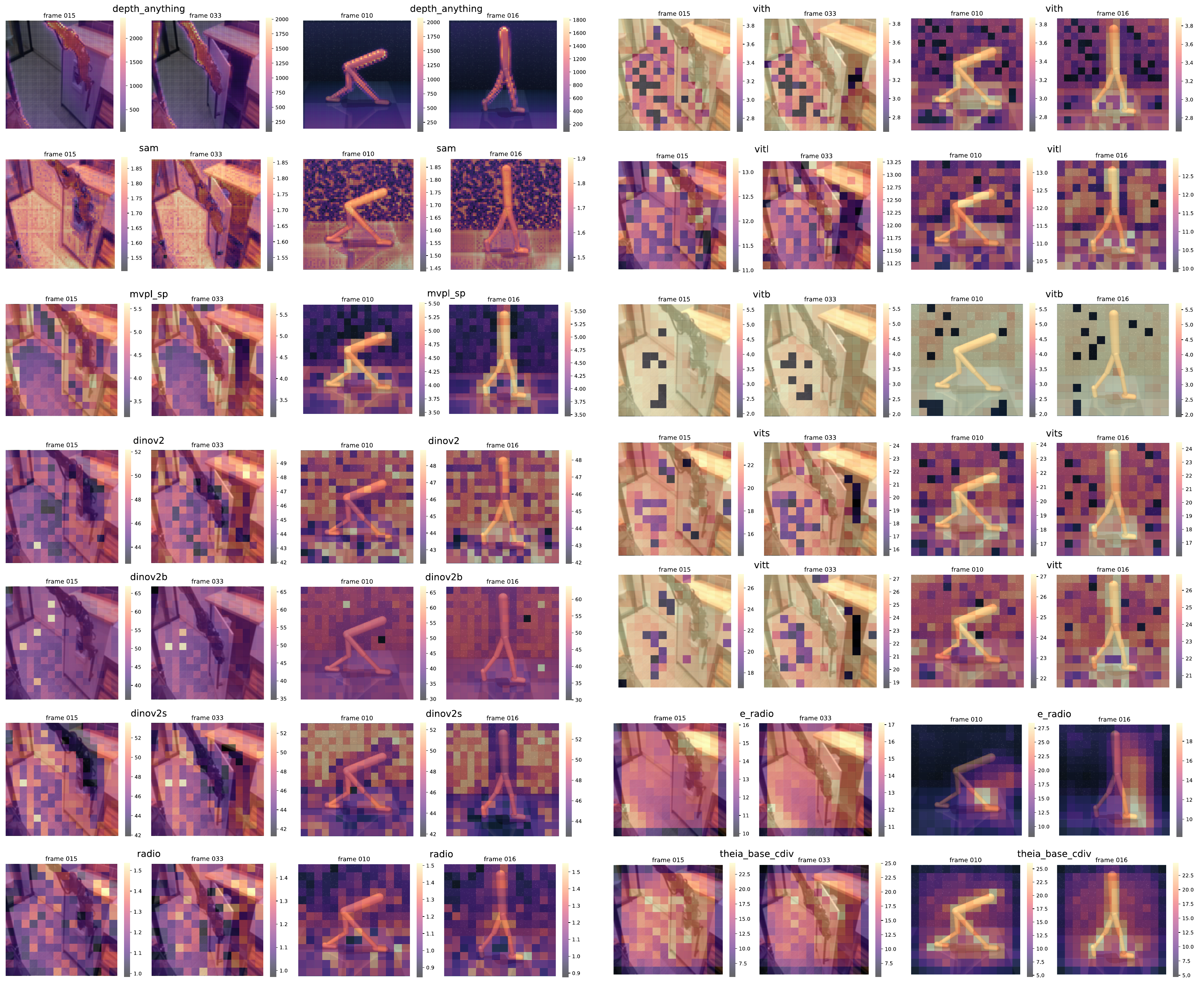}
    \caption{Feature norm map visualizations of ViT~\cite{dosovitskiy2021an}, DINOv2~\cite{oquab2023dinov2}, MVP~\cite{xiao2022masked}, Depth-Anything~\cite{yang2024depth}, SAM~\cite{kirillov2023segment}, RADIO~\cite{ranzinger2023amradio}, and E-RADIO~\cite{ranzinger2023amradio}.}
    \label{fig:feature_norm_map_all_other_models}
\end{figure}

%% file: figures/feature_norm_distribution.tex
\begin{figure}[thbp]
    \centering
    \includegraphics[width=\linewidth]{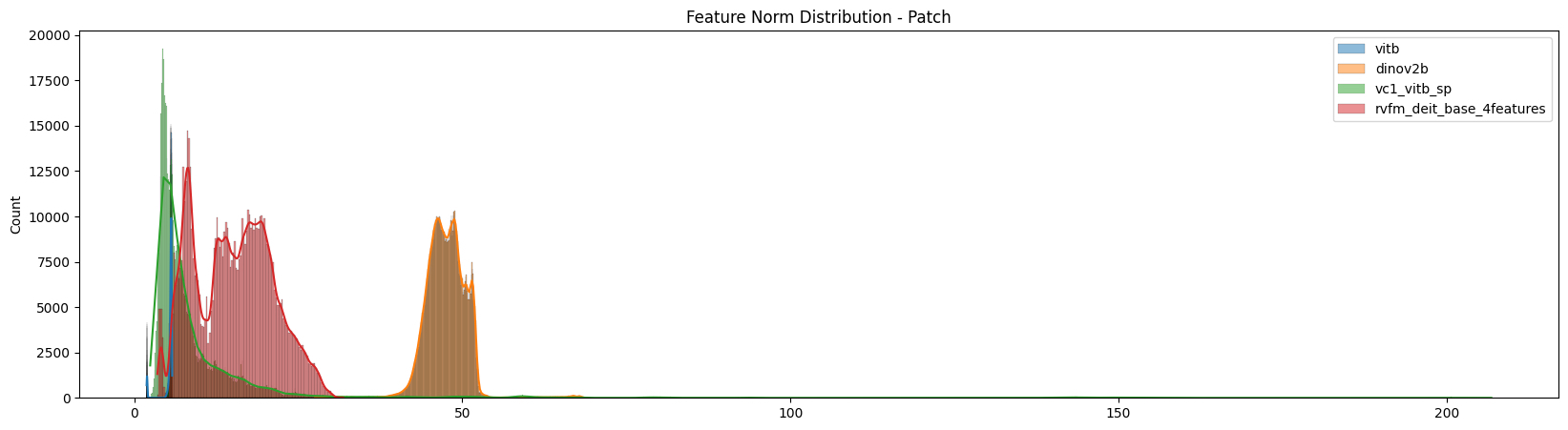}
    \includegraphics[width=\linewidth]{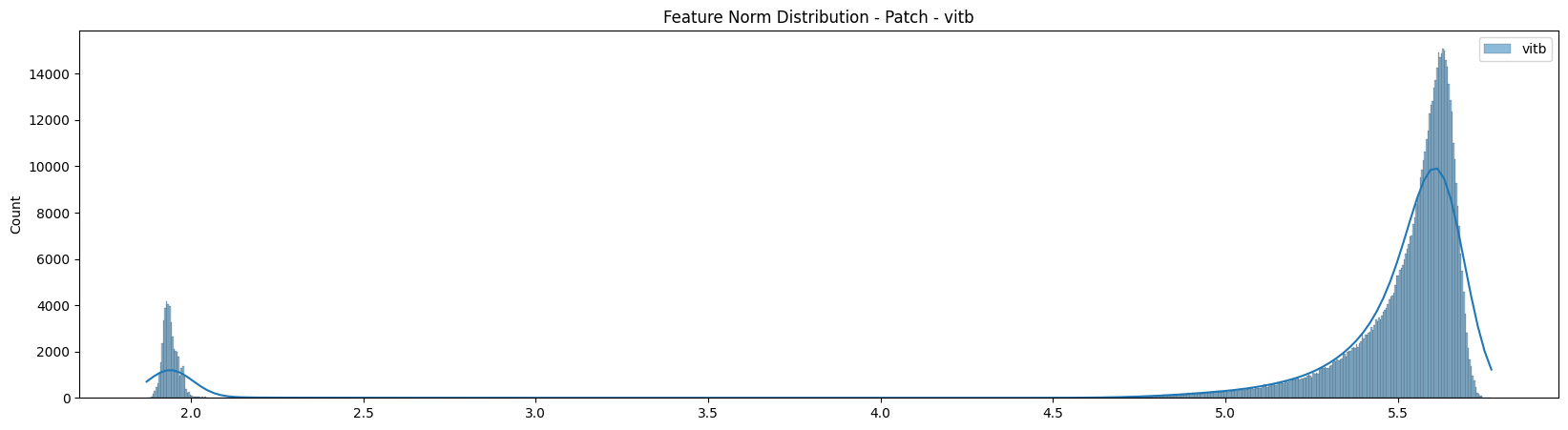}
    \includegraphics[width=\linewidth]{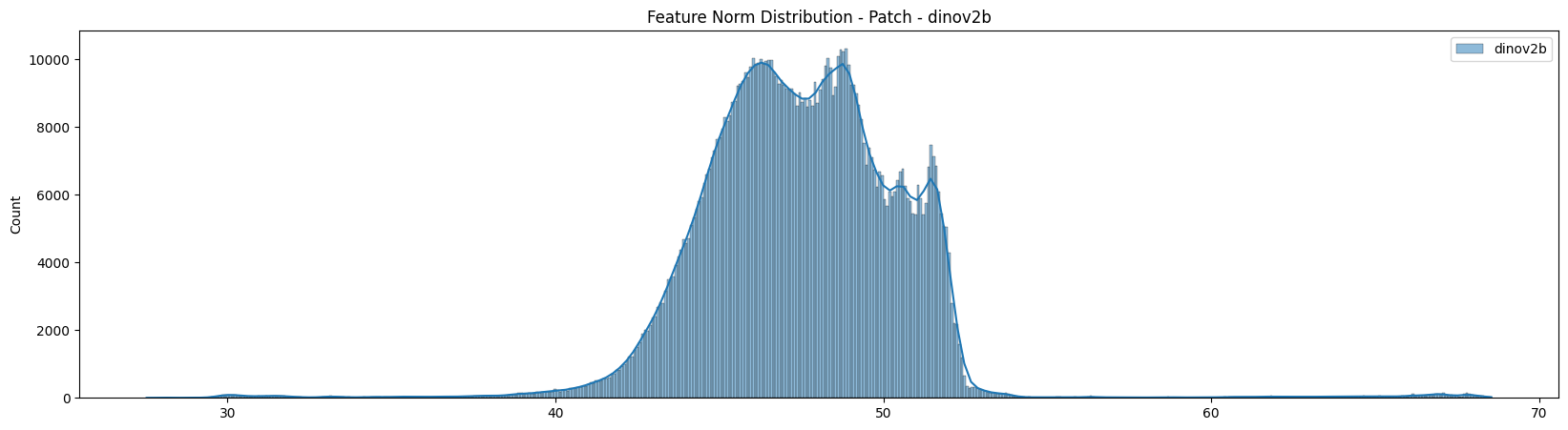}
    \includegraphics[width=\linewidth]{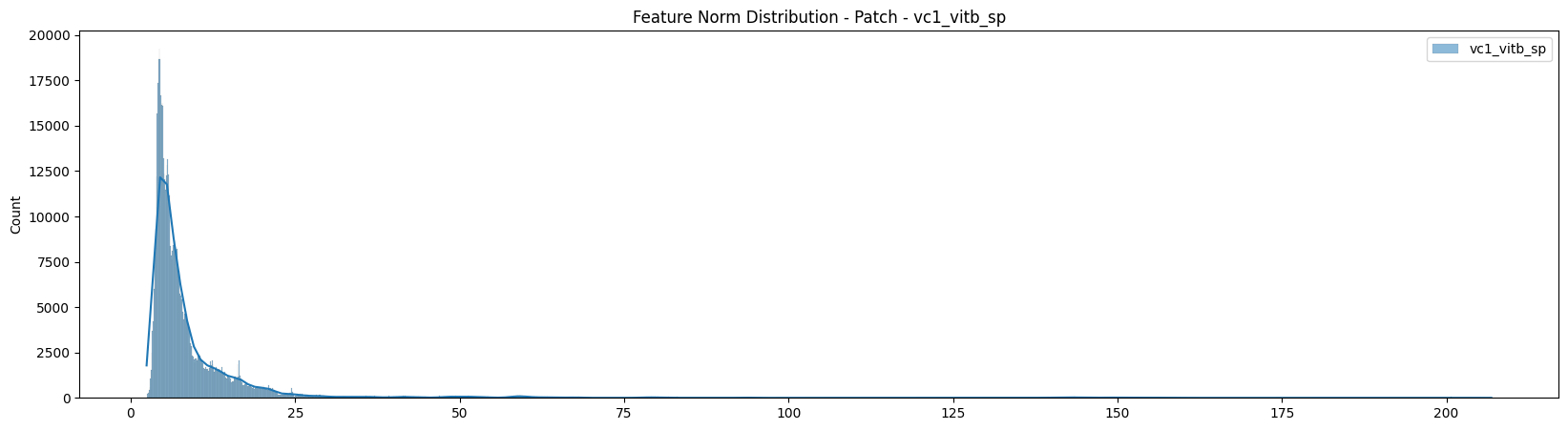}
    \includegraphics[width=\linewidth]{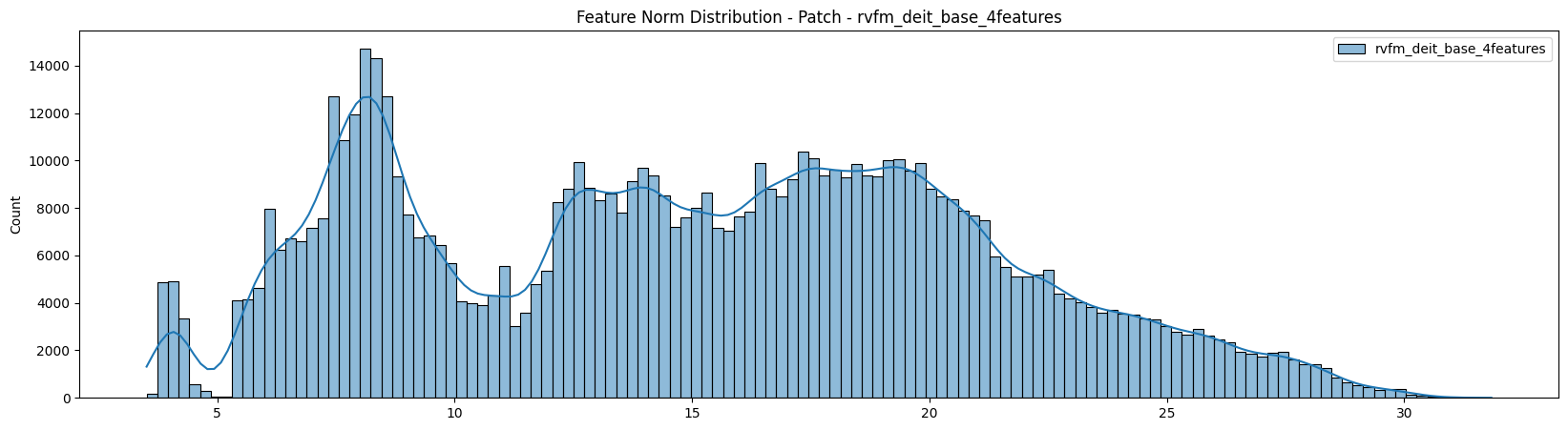}
    \caption{Distrubtions of feature norms. From top to bottom: 4 models on the same plot, ViT-B, DINOv2-B, VC-1-B, and Theia-B.}
    \label{fig:feature_norm_distribution}
\end{figure}

%% file: figures/feature_pcaauc_less.tex
\begin{figure}
    \centering
    \includegraphics[width=0.7\textwidth]{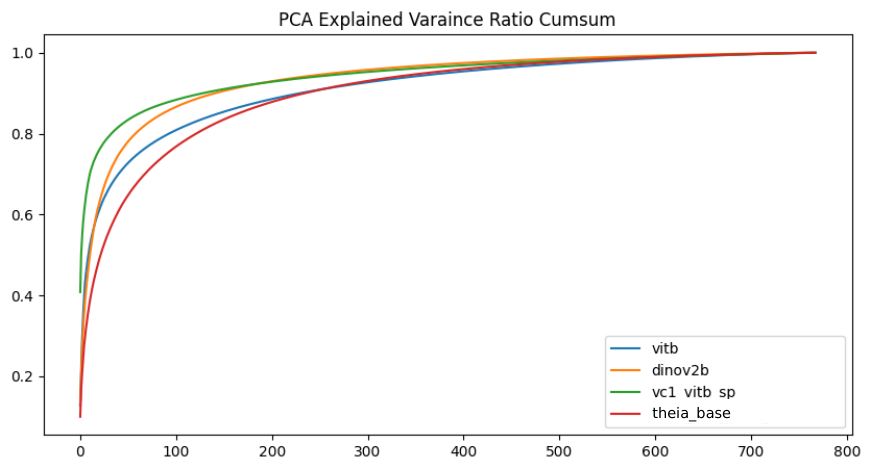}
    \caption{Cumulative sum of PCA Explained Variance Ratio of features from ViT-B, DINOv2-B, VC-1-B, and Theia-B.}
    \label{fig:feature_pca_evr_less}
\end{figure}

%% file: figures/feature_cossim_pcaauc_performance.tex
\begin{figure}
    \centering
    \includegraphics[width=0.504\textwidth]{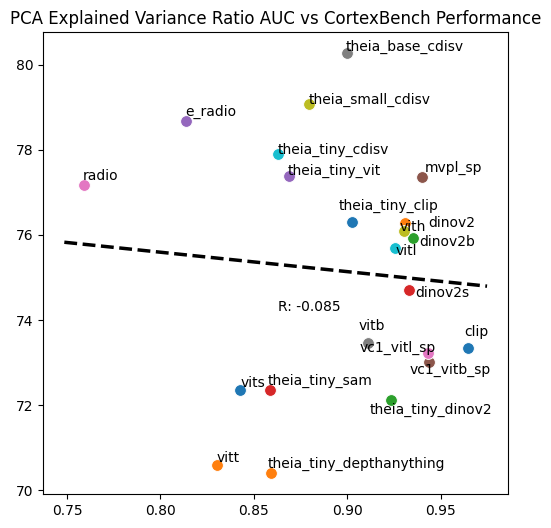}
    \includegraphics[width=0.47\textwidth]{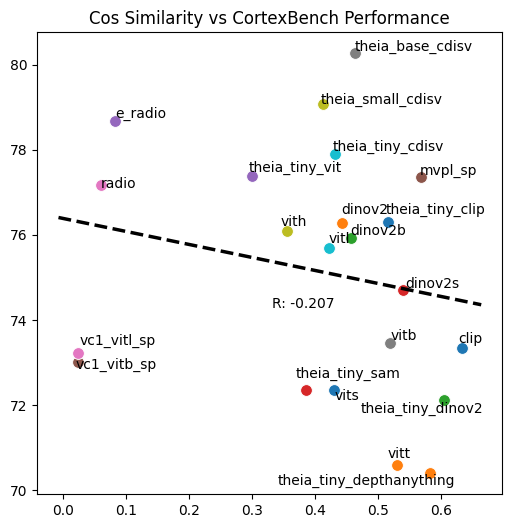}
    
    \caption{PCA explained variance ratio-AUC (left) and cosine similarity (right) vs MuJoCo performance of many models evaluated.}
    \label{fig:feature_cossim_pcaauc_performance}
\end{figure}

%% file: tables/imagenet_linear_probing.tex
\begin{table}[th]
    \centering
    \caption{ImageNet-1k~\cite{jia2009imagenet} evaluation accuracy using linear probing.}
    \begin{tabular}{cc}
    \toprule
        Model & Accuracy  \\
        \midrule
        Theia-B & 72.1\% \\
        Theia-B (initialized from DeiT-B~\cite{touvron2021training} weights) & 75.2\% \\
        MAE (ViT-B)~\cite{he2022masked} & 67.5\% \\
        DINOv2~\cite{oquab2023dinov2} (ViT-L) & 84.5\% \\
    \bottomrule
    \end{tabular}
    \label{tab:imagenet_linear_probing}
\end{table}

%% file: tables/probe3d.tex
\begin{minipage}{\textwidth}
    \begin{minipage}{0.32\textwidth}
        \centering
        \captionof{table}{Depth Estimation on NYU dataset}
        \resizebox{\textwidth}{!}{
        \begin{tabular}{lcccc}
        \toprule
        Model  & $\delta_1$ & $\delta_2$ & $\delta_3$ & RMSE \\
        \midrule
        Theia & 0.8420 & 0.9699 & 0.9929 & 0.4465 \\
        \bottomrule
        \end{tabular}
        }
        \label{tab:depth_estimation}
    \end{minipage}
    \begin{minipage}{0.32\textwidth}
        \centering
        \captionof{table}{ScanNet Multi-view Correspondence}
        \resizebox{\linewidth}{!}{
        \begin{tabular}{lcccc}
        \toprule
        Model  & $\theta_0^{15}$ & $\theta_{15}^{30}$ & $\theta_{30}^{60}$ & $\theta_{60}^{180}$ \\
        \midrule
        Theia & 47.88 & 37.47 & 24.78 & 13.18 \\
        \bottomrule
        \end{tabular}
        }
        \label{tab:scannet_multiview_correspondence}
    \end{minipage}
    \begin{minipage}{0.32\textwidth}
        \centering
        \captionof{table}{Surface Normal Estimation on NYU dataset}
        \resizebox{\textwidth}{!}{
        \begin{tabular}{lcccc}
        \toprule
        Model  & $11.25^{\circ}$ & $22.5^{\circ}$ & $30^{\circ}$ & RMSE \\
        \midrule
        Theia & 30.22 & 54.22 & 65.13 & 33.59 \\
        \bottomrule
        \end{tabular}
        }
        \label{tab:surface_normal_estimation}
    \end{minipage}
\end{minipage}

%% file: tables/generalization_fw.tex
\begin{table}[th]
    \centering
    \caption{Generalization evaluation on Factor-World.}
    \begin{tabular}{lcccccc}
    \toprule
     Model & Light & Table Texture & Table Pose & Camera Pose & Floor Texture & Arm Pose \\
     \toprule
     Theia-B & 100\% & 90\% & 100\% & 90\% & 100\% & 55\% \\
     E-RADIO & 100\% & 85\% & 80\% & 90\% & 100\% & 40\% \\
     MVP-L & 100\% & 50\% & 20\% & 70\% & 100\% & 45\% \\
     R3M & 100\% & 35\% & 70\% & 90\% & 100\% & 50\% \\
     \bottomrule
    \end{tabular}
    \label{tab:generlization_factor_world}
\end{table}